\documentclass[letterpaper, 10 pt, conference]{ieeeconf}  
\usepackage[style=ieee]{biblatex}       
\AtBeginBibliography{\footnotesize}
\usepackage{algpseudocode}              
\usepackage{algorithm}                  
\usepackage{ifthen}
\usepackage{graphicx}
\let\labelindent\relax
\usepackage{enumitem}                   
\usepackage{caption}
\usepackage{subcaption}                 
\usepackage{amsmath}
\usepackage{amssymb}
\usepackage{bbm}
\usepackage{amsfonts}
\usepackage{hyperref}                   
\usepackage{array}                      
\usepackage{multirow}                   
\usepackage{xcolor}
\usepackage{makecell}
\usepackage{changes}
\usepackage{acronym}
\newacro{MCQ}[MCQ]{multiple-choice question}
\newacro{BB}[BB]{bounding box}
\newacro{OE}[OE]{open-ended}

\usepackage{multicol}

\newboolean{isarxiv}
\setboolean{isarxiv}{true}

\newcommand{\appendixcite}[1]{%
    \ifthenelse{\boolean{isarxiv}}%
        {}%
        { \cite{#1}}%
}

\IEEEoverridecommandlockouts                              

\title{\LARGE \bf
To Ask or Not To Ask: Human-in-the-loop Contextual Bandits \\ with Applications in Robot-Assisted Feeding
}

\author{Rohan Banerjee$^{1}$, Rajat Kumar Jenamani$^{*1}$, Sidharth Vasudev$^{*1}$, \\ Amal Nanavati$^{2}$, Katherine Dimitropoulou$^{3}$, Sarah Dean\textsuperscript{\textdagger}$^{1}$, Tapomayukh Bhattacharjee\textsuperscript{\textdagger}$^{1}$
\thanks{*Equal Contribution, \textsuperscript{\textdagger}Equal Advising. This work was partly funded by NSF CCF 2312774 and NSF OAC-2311521, a LinkedIn Research Award, a
gift from Wayfair, NSF IIS \#2132846, CAREER \#2238792, and NIH \#T32HD113301. Full acknowledgements in Appendix.}
\thanks{$^{1}$Department of Computer Science, Cornell University  {\tt\small \{rbb242, rj277, sv355, sdean, tapomayukh\}@cornell.edu}}%
\thanks{$^{2}$Department of Computer Science and
Engineering, University of Washington, Seattle, WA 98195 {\tt\small \{amaln\}@cs.washington.edu}}%
\thanks{$^{3}$Columbia University, New York
City, NY, USA,  {\tt\small kd2524@cumc.columbia.edu}}%
}

\DeclareCiteCommand{\citeconsecutive}
  {\usebibmacro{prenote}}
  {\usebibmacro{citeindex}%
   \printtext[bibhyperref]{\printfield{labelnumber}}}
  {\multicitedelim}
  {\usebibmacro{postnote}}

\renewcommand{\baselinestretch}{0.98}   

\begin{document}

\maketitle
\thispagestyle{empty}
\pagestyle{empty}

\begin{abstract}

Robot-assisted bite acquisition involves picking up food items with varying shapes, compliance, sizes, and textures. Fully autonomous strategies may not generalize efficiently across this diversity. We propose leveraging feedback from the care recipient when encountering novel food items. However, frequent queries impose a workload on the user. We formulate human-in-the-loop bite acquisition within a contextual bandit framework and introduce \textsc{LinUCB-QG}, a method that selectively asks for help using a predictive model of querying workload based on query types and timings. This model is trained on data collected in an online study involving 14 participants with mobility limitations, 3 occupational therapists simulating physical limitations, and 89 participants without limitations. We demonstrate that our method better balances task performance and querying workload compared to autonomous and always-querying baselines and adjusts its querying behavior to account for higher workload in users with mobility limitations. We validate this through experiments in a simulated food dataset and a user study with 19 participants, including one with severe mobility limitations. Please check out our project website at: \href{https://emprise.cs.cornell.edu/hilbiteacquisition/}{\textbf{emprise.cs.cornell.edu/hilbiteacquisition/}}.

\end{abstract}

\section{INTRODUCTION}


Feeding, an essential Activity of Daily Living (ADL) \cite{katz1963studies}, is challenging for individuals with mobility limitations, often requiring caregiver support that may not be readily available. Approximately 1 million people in the U.S. cannot eat without assistance \cite{brault2012americans}, so robotic systems could empower these individuals to feed themselves, promoting independence. Robot-assisted feeding comprises two key tasks \cite{madan2022sparcs}: bite acquisition \cite{feng2019robot, gordon2020adaptive, gordon2021leveraging, sundaresan2022learning}, the task of picking up food items, and bite transfer [\citeconsecutive{gallenberger2019transfer, belkhale2022balancing}], delivering them to the user's mouth. Our work focuses on bite acquisition, aiming to learn policies that robustly acquire novel food items with diverse properties.

With over 40,000 unique food items in a typical grocery store \cite{malito2017grocery}, bite acquisition strategies must generalize well to novel food items. Current approaches formulate bite acquisition as a  contextual bandit problem [\citeconsecutive{gordon2020adaptive,gordon2021leveraging}], using online learning techniques like \textsc{LinUCB} \cite{li2010contextual} to adapt to unseen food items. However, these methods may need many samples to learn the optimal action, which is problematic due to food fragility and potential user dissatisfaction from failures \cite{bhattacharjee2020more}. 

Our key insight is that we can leverage the care recipient's presence to develop querying strategies that can adapt to novel foods, giving users a sense of agency. Under the assumption that the human can specify the robot's acquisition action, we extend the contextual bandit formulation to include human-in-the-loop querying \cite{cui2021understanding}. However, excessive querying may overwhelm users and reduce acceptance \cite{fong2003robot}. Thus, our key research question is: How do we balance imposing \textit{minimal querying workload} on the user while achieving \textit{maximal bite acquisition success} in a human-in-the-loop contextual bandit framework? 

\begin{figure}[t]
  \centering
    \includegraphics[width=\linewidth]{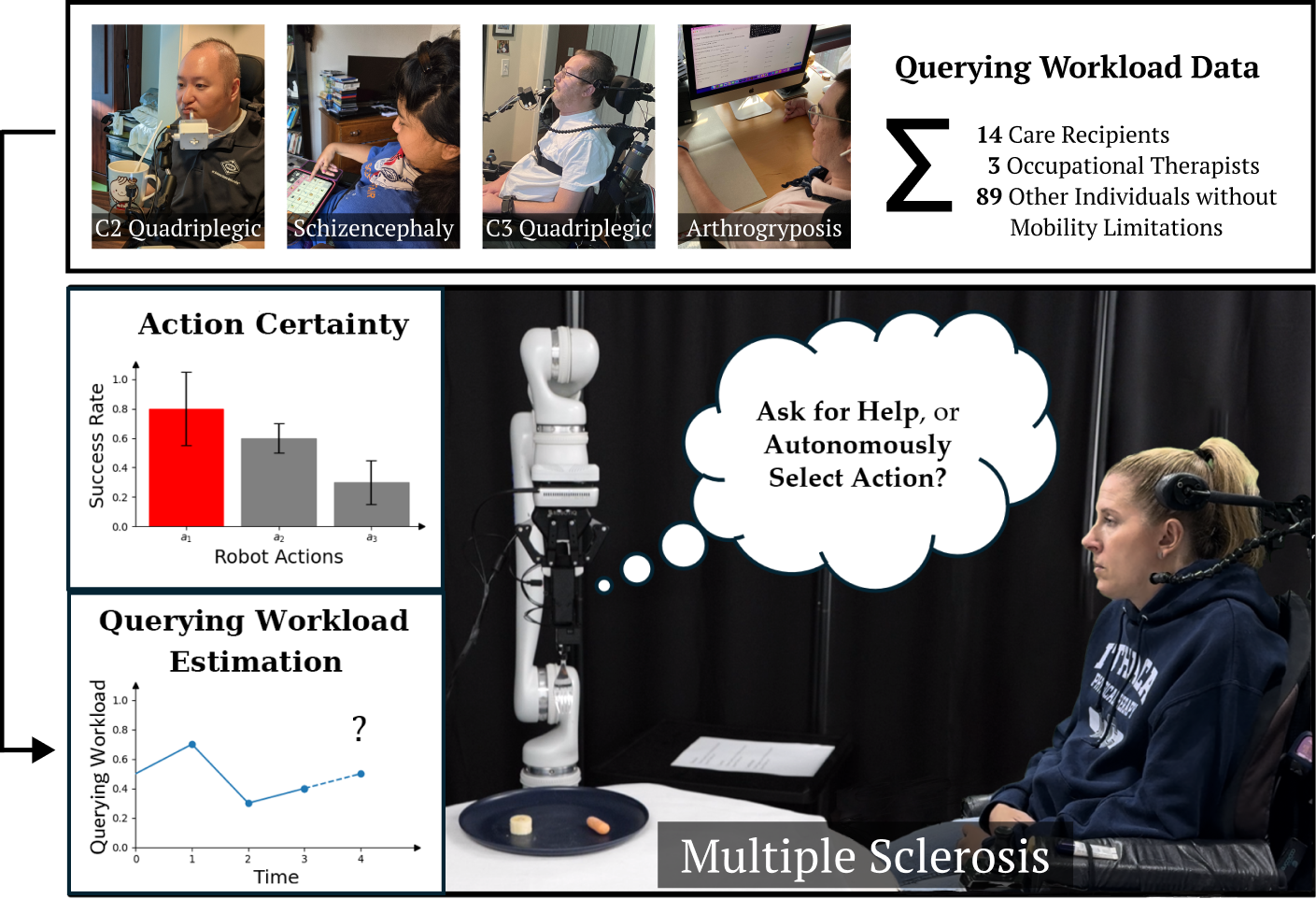}
    \captionsetup{font=small}
    \caption{We present a human-in-the-loop contextual bandit-based framework for robot-assisted bite acquisition and propose a novel method which decides whether to ask for help or autonomously act, by considering both certainty about action performance, and the estimated workload imposed on the human by querying.}
    \label{fig:fig1}
\end{figure}

\begin{figure*}[t]
  \centering
  \vspace{0.2cm}
    \includegraphics[width=0.9\textwidth]{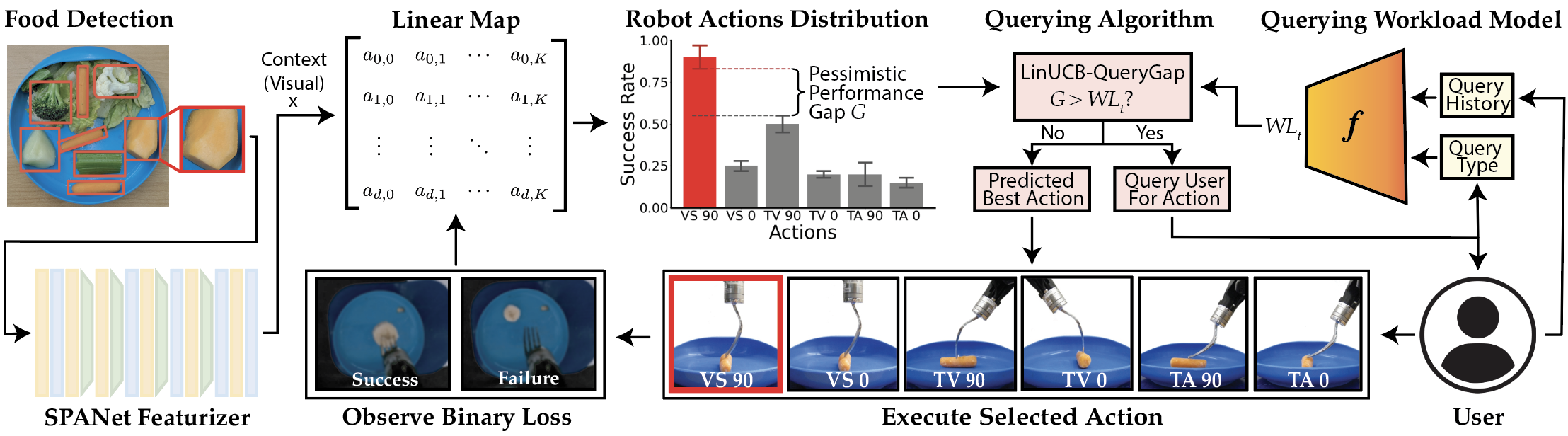}
    \captionsetup{font=small}
    \caption{Human-in-the-loop contextual bandit pipeline for deciding whether to query ($a_q$) or autonomously select a robot action $a_r$. Our proposed method, \textsc{LinUCB-QG}, takes into account action uncertainty as measured by the performance gap $G$ between the best and the second best robot action and incorporates a learned querying workload model ($f$) to predict the workload of querying the human $WL_{t}$.}
    \vspace{-0.45cm}
    \label{fig:hil-cb-pipeline}
\end{figure*}

Balancing querying workload requires accurately estimating the workload imposed on users. Existing methods rely on physiological sensors \cite{shayesteh2021investigating, aygun2022investigating, fridman2018cognitive}, which can be invasive. We develop a non-intrusive, data-driven method by conducting a study on 14 participants with mobility limitations, 3 occupational therapists (OTs) who simulate the physical limitations of those with mobility limitations, and 89 participants without mobility limitations. We gather data on how different query types and timings impact self-reported workload using a modified NASA-TLX scale \cite{hart1988development}. We tailor queries to common autonomous system failures, ranging from multiple-choice questions to open-ended prompts for acquisition strategy suggestions.

Using this data, we train a model to predict user workload in response to queries, considering the nature of the query and the user's prior query interactions with the system. We observe statistically significant differences in self-reported workload between users with and without mobility limitations, which are accounted for by our learned models.

Building on our workload model, we propose \textsc{LinUCB-QueryGap} (\textsc{LinUCB-QG}), a novel algorithm for human-in-the-loop contextual bandits. It decides when to query based on the estimated querying workload and performance uncertainty of candidate actions. Simulated experiments on a dataset of 16 food items \cite{feng2019robot} show that \textsc{LinUCB-QG} is able to better balance workload and bite acquisition success compared to three baselines: (i) \textsc{LinUCB} \cite{gordon2020adaptive}, a state-of-the-art fully autonomous approach; (ii) \textsc{AlwaysQuery}, which always queries; and (iii) \textsc{LinUCB-ExpDecay}, a naive querying algorithm. \textsc{LinUCB-QG} achieves higher task performance than \textsc{LinUCB} and \textsc{LinUCB-ExpDecay}, with lower workload than \textsc{AlwaysQuery}. It also queries less when using a workload model trained on users with mobility limitations, indicating sensitivity to the workload differences across diverse user populations.

We validate our method in a real-world user study with 19 individuals, including one with Multiple Sclerosis, using three foods with variable material properties: banana slices, baby carrots, and cantaloupe. \textsc{LinUCB-QG} achieves a statistically significant 26\% higher task success compared to \textsc{LinUCB}, and a 47\% lower change in querying workload compared to \textsc{AlwaysQuery}.

Our contributions are as follows:
\begin{itemize}[leftmargin=*,itemsep=-0.1em]
    \item A human-in-the-loop contextual bandit framework incorporating querying workload for bite acquisition.
    \item A dataset (including users with mobility limitations) on how feeding-related queries affect workload, and a predictive workload model without exteroceptive sensor inputs.
    \item A novel method, \textsc{LinUCB-QG}, that balances acquisition success and querying frequency using our workload model, shown through simulations and a real-world study with 19 users, including one with severe mobility limitations.
\end{itemize}


\section{RELATED WORK}
\vspace{-0.15cm}

\textbf{Human-in-the-loop algorithms.} Our problem is an instance of learning to defer to an expert \cite{raghu2019algorithmic}, where an agent decides whether to act autonomously or defer to experts like humans or oracle models. Most approaches operate in supervised learning \cite{raghu2019algorithmic, keswani2021towards, narasimhan2022post, mozannar2020consistent} or reinforcement learning domains \cite{joshi2021learning}. Given the constraints of robot-assisted bite acquisition (observing independent contexts and receiving sparse feedback) we focus on learning-to-defer policies in an online contextual bandit setting. While prior methods often assume a fixed cost for querying the expert \cite{keswani2021towards, narasimhan2022post, mozannar2020consistent} or deal with imperfect experts \cite{narasimhan2022post}, we propose estimating a time-varying deferral penalty using a data-driven model.

Other research explores active learning in robotics \cite{fitzgerald2022inquire, racca2019teacher, li2023embodied}, where agents decide which instances to query [\citeconsecutive{racca2019teacher, li2023embodied}] or which type of feedback to solicit \cite{fitzgerald2022inquire}. Some algorithms balance task performance and human querying cost, but our work differs by using a data-driven model of querying cost and applying it to the complex task of bite acquisition.

\textbf{Querying workload modeling.} Our algorithms need to monitor workload to avoid over-querying users. For individuals with mobility limitations, querying workload may involve physical \cite{smith2022decomposed} and cognitive components \cite{hart1988development}, depending on the query type. Most workload estimation literature relies on neurological/physiological signals (ECG/heart rate, respiration \cite{smith2022decomposed}, physical posture \cite{harriott2013assessing} for physical workload; EEG \cite{shayesteh2021investigating}, EDA \cite{rajavenkatanarayanan2020towards}, pupil metrics [\citeconsecutive{ahmad2019trust, fridman2018cognitive}] for cognitive workload), training supervised models based on these signals. However, measuring such signals requires specialized, often invasive equipment. Alternatively, workload can be estimated post-hoc using subjective metrics like NASA-TLX \cite{shayesteh2021investigating}.

In contrast, we develop a data-driven predictive model of querying workload using self-reported, modified NASA-TLX survey results, without relying on specialized sensors. Our model estimates the expected workload of a query based on interaction history, timing, and query type. Additionally, unlike prior work [\citeconsecutive{shayesteh2021investigating, fridman2018cognitive, rajavenkatanarayanan2020towards, ahmad2019trust}], we focus on workload estimation for users with mobility limitations.

\vspace{-0.1cm}
\section{PROBLEM FORMULATION}
\label{sec:problem-formulation}
\vspace{-0.15cm}

Following previous work [\citeconsecutive{gordon2020adaptive, gordon2021leveraging}], we formulate bite acquisition as a contextual bandit problem. At each timestep, the learner receives a context $x_t$ (food item), and selects actions $a_t \in \mathcal{A}$ that minimize regret over $T$ timesteps:

\vspace{-0.15cm}

\begin{equation*}
    \sum_{t=1}^T \mathbb{E}[r_t | x_t, a^*_t, t] - \sum_{t=1}^T \mathbb{E}[r_t | x_t, a_t, t]
\end{equation*}

\vspace{-0.15cm}

where $r_t$ is the reward, $a^*_t \triangleq a^*(x_t)$ is the optimal action maximizing $r_t$ for $x_t$, and expectations are over the stochasticity in $r_t$. In our problem setting, we assume access to a dataset $D$ from a robotic manipulator, consisting of observations $o \in \mathcal{O}$, actions $a \in \mathcal{A}$, and rewards $r$. This dataset is used to pretrain and validate our algorithms, but is not required for our online setting. Our contextual bandit setting is characterized by (more details in Appendix\appendixcite{appendix}):

\begin{itemize}[leftmargin=*,itemsep=-0.1em]
    \item \textbf{Observation space $\mathcal{O}$}: RGB images of single bite-sized food items on a plate, sampled from 16 food types \cite{feng2019robot}.
    \item \textbf{Action space $\mathcal{A}$}: 7 actions in total: 6 robot actions $a_r$ shown in Fig. \ref{fig:hil-cb-pipeline} (bottom) [\citeconsecutive{bhattacharjee2019towards, feng2019robot}], and 1 query action $a_q$.
    \item \textbf{Context space $\mathcal{X}$}: Lower-dimensional context $x$ derived from observations $o$ using SPANet \cite{feng2019robot}, with dimensionality $d=2048$. As our bandit algorithms assume a linear relationship between $x$ and the expected reward $\mathbb{E}[r]$, we use the penultimate activations of SPANet as our context since the final layer of SPANet is linear \cite{gordon2020adaptive}.    
\end{itemize}

When the learner selects a robot action $a_r$, it receives a binary reward indicating success for context $x_t$. If the learner selects $a_q$, it receives the optimal robot action $a^*(x_t)$ from the human. It then executes that action and receives a reward. Querying imposes a penalty on the learner, given by the human's latent querying workload state $WL_t$ at time $t$.

We model the total expected reward $\mathbb{E}[r_t|x_t, a_t, t]$ as the sum of the expected task reward $r_{task}(x_t,a_t)$ (the probability that action $a_t$ succeeds for food context $x_t$) and the expected querying workload reward $r_{WL}(x_t,a_t,t)$ associated with selecting query actions $a_q$:

\vspace{-0.5cm}

\begin{equation*}
    \mathbb{E}[r_t|x_t, a_t, t] \triangleq w_{task} r_{task}(x_t,a_t) + (1-w_{task}) r_{WL}(x_t,a_t,t)
\end{equation*}
where $w_{task}$ is a scalar weight in $[0,1]$. We define $r_{task}(x_t,a_t)$ to be $s_x(x_t,a_t)$ if $a_t = a_r$, and $s_x(x_t, a^*(x_t))$ if $a_t = a_q$, where $s_x(x_t,a_t)$ is the probability that action $a$ succeeds for context $x$. We define $r_{WL}(x_t,a_t,t)$ to be $0$ if $a_t=a_r$, and $-WL_t$ if $a_t=a_q$. 
Our goal is to learn a policy $\pi(a|x,t)$ that maximizes $ \sum_{t=1}^T \mathbb{E}[r_t|x_t, a_t, t]$.

\begin{figure*}[t]
    \centering
    \vspace{0.1cm}
    \includegraphics[width=\textwidth]{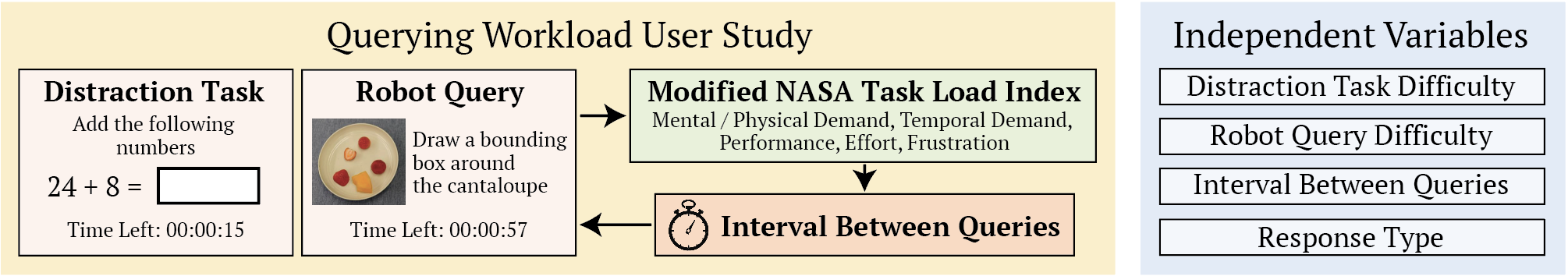}
     \vspace{-0.1cm}
    \captionsetup{font=small}
    \caption{\textit{Left:} Querying workload user study setup, illustrating the format of the distraction task and robot query task, and the modified NASA-TLX survey recorded after every robot query. \textit{Right:} Independent variables that affect user's querying workload varied in the study. }
    \vspace{-0.45cm}
    \label{fig:prolific-study}
\end{figure*}

\section{querying workload Modeling}

Our human-in-the-loop contextual bandit algorithms rely on estimating the querying workload state $WL_t$ (Sec. \ref{sec:problem-formulation}). To estimate $WL_t$, we first conduct a user study to understand how workload responds to queries (Sec. \ref{subsec:user-study}). We then learn a predictive workload model using this data (Sec. \ref{subsec:predictive-model}).

\subsection{Querying Workload User Study}
\label{subsec:user-study}

We design a user study (Fig. \ref{fig:prolific-study}) to capture how different query types affect an individual's querying workload in robotic caregiving scenarios. In such scenarios, a feeding system may request various assistance types from the user, such as semantic labels, bounding boxes for food items, or explanations for failed acquisition attempts. These response types can have varying impacts on workload [\citeconsecutive{cui2021understanding, koppol2021interaction}].

For instance, a semantic label response may impose less workload than an open-ended response. In addition, the end-user of the system and the caregiver may be busy with other tasks, such as watching TV while eating. We conduct an online user study that captures the above factors to determine how an individual's querying workload changes in response to requests for assistance from an autonomous system. 

Participants perform a distraction task while periodically receiving queries from a ``robot," simulating realistic caregiving situations where users might be engaged in other activities. The robot queries represent realistic assistance requests during bite acquisition. We vary four independent variables hypothesized to impact querying workload:
\begin{enumerate}
    \item Robot query difficulty ($d_t$): 2 options -- easy or hard.
    \item Distraction task difficulty ($dist_t$): 3 options -- no distraction task, an easy numeric addition task, or a hard video transcription task.
    \item Time interval between queries ($\Delta T$): 2 options -- either 1 minute or 2 minutes.
    \item Response type ($resp_t$): 3 options -- (a) a \ac{MCQ} asking for a semantic label, (b) asking the user to draw a \ac{BB} around a food item, or (c) an \ac{OE} question asking the user to explain why an acquisition attempt failed. 
\end{enumerate}

Each participant experiences 12 study conditions, one for each of the 2 settings of $d_t$, 3 settings of $dist_t$, and 2 settings of $\Delta T$, with the entire study taking roughly 1.5 hr. During each condition, lasting 5.5 minutes, participants engage in distraction tasks and respond to robot queries (with varying response types $resp_t$) as shown in Fig. \ref{fig:prolific-study}. We measure self-reported workload (our dependent variable) after each robot query and at the end of each condition using 5 modified NASA-TLX subscales [\citeconsecutive{hart1988development, hertzum2021reference}] (mental/physical, temporal, performance, effort, frustration), each measured on a 5-point Likert scale (see Appendix for exact question wording\appendixcite{appendix}).

We collect data from 14 users (7 male, 7 female; ages 27-50) with mobility limitations resulting from a diverse range of medical conditions: spinal muscular atrophy (SMA), quadriplegia, arthrogryposis, cerebral palsy, Ullrich congenital muscular dystrophy,  schizencephaly with spastic quadriplegia, spinal cord injury (including C2 and C3 quadriplegia). We additionally collect data from 3 occupational therapists (OTs), (3 female; ages 23-26), trained to simulate right/left hemiplegia (stroke). The OTs positioned one side of their body in full shoulder adduction, and elbow flexion to allow only 30 degrees of functional elbow joint range. Finally, we collect data from 89 users without mobility limitations (40 male, 49 female; ages 19-68). The population size for users with limitations is larger than the median ($n=6$) in studies of assistive robots [\citeconsecutive{mankoff2010disability, nanavati2023physically}].


\subsection{Data Analysis}
\label{subsec:workload-data-analysis}

We consider three distinct datasets under which to train the workload model: $D_1$, which only includes the data from the 89 users without mobility limitations; $D_2$, which only includes the data from the 14 users with mobility limitations and 3 OTs; and $D_{1,2}$, which includes both $D_1$ and $D_2$. We convert the modified NASA-TLX responses into a single querying workload score by normalizing responses to 3 of the questions, and taking a weighted average (weight details in Appendix\appendixcite{appendix}). We find a statistically significant difference in mean querying workload between $D_1$ ($0.36 \pm 0.29$) and $D_2$ ($0.40 \pm 0.28$), as described in the Appendix\appendixcite{appendix}.

\subsection{Querying Workload Predictive Model}
\label{subsec:predictive-model}

We develop a predictive model of querying workload based on the user study data (Sec. \ref{subsec:user-study}). Our models operate in discrete time, where the time variable $t$ refers to an integer number of timesteps since the beginning of a condition, where each timestep corresponds to a fixed time spacing $\Delta t$ (set to $10 \text{s}$ in our experiments). The model has the form $WL_{t} = f(WL_0, Q_t; \theta)$, where $WL_0$ is the initial workload, $Q_t = \{(d_{t'}, resp_{t'},  dist_{t'})\}_{t'=1}^{t}$ is the history of previous queries (where $d_t$, $resp_t$, and $dist_t$ are the query variables defined in Sec. \ref{subsec:user-study}), and $\theta$ are the model parameters. 

\textbf{Linear discrete-time models.} To capture the dependency between the $WL_t$ and $Q_t$, we use a linear predictive workload model inspired by Granger causality models [\citeconsecutive{granger1969investigating,liang2023randomization}]:
\begin{equation*}
     WL_{t} = \gamma WL_{0} + \sum_{i=0}^{H-1} \mathbf{w_i}^T [d_{t-i}, resp_{t-i}, dist_{t-i}] + w_0
\end{equation*}

\vspace{-0.15cm}

where $\mathbf{w_i}$ represents the effect of the query asked $i$ timesteps in the past on the workload at the current time $t$, $H$ is the history length, and $w_0$ is a bias term. We train the model using linear regression, where we convert $d_t$, $resp_t$, and $dist_t$ into features using one-hot encodings, and generate a training pair for each robot task query in our study (see Appendix\appendixcite{appendix} for more details about the Granger models).

\textbf{Model selection.} We train models $f_1$, $f_2$, and $f_{1,2}$, corresponding to the datasets $D_1$, $D_2$, and $D_{1,2}$, respectively (Sec. \ref{subsec:workload-data-analysis}). \footnote{When training these models, our assumptions about the time taken for users to complete the modified TLX surveys, particularly for users without mobility limitations, are in the Appendix.\appendixcite{appendix}}. The models $f_{i,sim}$ and $f_{i,real}$ indicate models learned for the simulation experiments and the real-world user study, respectively. For our simulation experiments, we learn 3 separate models $f_{1,sim}$, $f_{2,sim}$, and $f_{1,2,sim}$ (details in the Appendix\appendixcite{appendix}). For our real-world user study, we only use 2 of the 3 real models: $f_{1,real}$ for the users without limitations and $f_{1,2,real}$ for the users with limitations. In our setting, $f_{1,real}$ and $f_{1,2,real}$ are models chosen based on cross-validation median test MSE on $D_1$ and $D_{1,2}$, respectively (full model selection details in Appendix\appendixcite{appendix}). We also train a model $f_{2,real}$ (cross-validation score: $0.066 \pm 0.022$), which has a higher mean and standard deviation MSE than $f_{1,2,real}$ (cross-validation score: $0.058 \pm 0.010$). This suggests that training only on $D_2$ leads to a higher-variance, less accurate model, compared to training on $D_{1,2}$. Thus, it is still helpful to consider the data from users without limitations ($D_1$) when training our workload models. 

\vspace{-0.1cm}

\section{Human-in-the-Loop Algorithms}
\label{sec:hil-algorithms}

We develop decision-making algorithms that decide whether to ask for help or act autonomously, using the learned workload models in Sec. \ref{subsec:predictive-model}. Specifically, we consider four human-in-the-loop contextual bandit algorithms: one fully autonomous, and three that can query the human.

\textbf{Fully autonomous algorithm.} Our fully autonomous baseline is LinUCB \cite{li2010contextual}, the state-of-the-art for acquiring unseen bite-sized food items like banana and apple slices [\citeconsecutive{gordon2020adaptive, gordon2021leveraging}]. LinUCB selects the action that maximizes a reward upper-confidence bound (UCB) estimate for each robot action, given by $UCB_{a_r} = \theta_{a_r}^T x + \alpha b_{a_r}$. Here, $\theta_{a_r}$ is the linear parameter vector learned through regression on contexts $X_{a_r}$ and observed rewards for actions $a_r$, where $X_{a_r}$ includes contexts seen for $a_r$ during pretraining and online validation. The term $\alpha > 0$ corresponds to a confidence level, and $b_{a_r} = (x^T (X_{a_r}^T X_{a_r} + \lambda I)^{-1} x)^{1/2}$ is the UCB bonus with $L_2$ regularization strength $\lambda$. The size of $\alpha b_{a_r}$ reflects the reward estimate uncertainty for the given context-action pair.

\vspace{-1ex}
\begin{algorithm}
\caption{\textsc{LinUCB-QG}.}
\label{alg:LinUCB-Query-Gap}
\small
\begin{algorithmic}[1]
\State Inputs: Context $x$, scaling factor $w$, time $t$, initial workload $WL_0$, workload model $f$, model parameters $\theta$
\State For all $a_r$: compute UCB bonus and value: ($b_{a_r}$, $UCB_{a_r}$).
\State Let $a^* = \arg \max_{a_r} \theta_{a_r}^T x$, $a^- = \arg \max_{a_r \neq a^*} \theta_{a_r}^T x$
\State Define $G \triangleq (\theta_{a^-}^T x + \alpha b_{a^-}) - (\theta_{a^*}^T x - \alpha b_{a^*})$
\State{Set $a = \begin{cases}
    \text{$a_q$} & \text{if $G > w f(WL_0, Q_t; \theta)$} \\
    \text{$\arg \max_{a_r} UCB_{a_r}$} & \text{otherwise}
\end{cases}$
}\\
\Return $a$
\end{algorithmic}
\end{algorithm}
\vspace{-1ex}
\textbf{Querying algorithms.} The querying algorithms decide to query the human or select the action that maximizes $UCB_{a_r}$.

\begin{enumerate}[leftmargin=*]
    \item \textsc{AlwaysQuery}: always queries the user. \footnote{For a particular food context $x$, we assume that the user provides the optimal action $a^*(x)$ when queried. However, the expert action may sometimes fail due to inherent action uncertainty (e.g. due to food property variability). Thus, if $a^*(x)$ initially fails, \textsc{AlwaysQuery} will repeatedly execute $a^*(x)$ until success.}
    \item \textsc{LinUCB-ExpDecay}: queries with exponentially-decaying probability (decay rate $c$) depending on number of food items seen in an episode (see Appendix\appendixcite{appendix}). 
    \item \textsc{LinUCB-Query-Gap} (\textsc{LinUCB-QG}), defined in Algorithm \ref{alg:LinUCB-Query-Gap}: queries if the worst-case performance gap between the best action $a^*$ and second-best action $a^{-}$ exceeds the predicted workload, with scaling factor $w$.
\end{enumerate}

\begin{table}[t]
\vspace{0.2cm}
\centering
\tiny
\begin{tabular}{ |p{1.0 in}||c|c|c| } 
\hline
\textbf{} & $r_{task,avg}$ & $M_{wt}$ & $f_q$ \\ 
 \hline
 \hline
Workload Model $f_{1,sim}$ &  &   & \\
 \hspace{0.1cm}   \textendash \textsc{LinUCB} & $0.292 \pm 0.113$ & $0.354 \pm 0.079$ & -\\

   \hspace{0.1cm}  \textendash \textsc{AlwaysQuery} & $0.727 \pm 0.195$ & $0.591 \pm 0.130$ & $1 \pm 0$ \\

   \cline{1-4}
 \hspace{0.1cm}   \textendash \textsc{LinUCB-ExpDecay} & $0.399 \pm 0.129$ & $0.425 \pm 0.090$ & $0.560 \pm 0.150$ \\

 \hspace{0.1cm}   \textendash \textsc{LinUCB-QG} & $0.670 \pm 0.237$ & $\mathbf{0.593 \pm 0.130}$ & $\mathbf{0.867 \pm 0.189}$ \\
 \hline
 \hline
 Workload Model $f_{2,sim}$ &  &  &    \\
 \hspace{0.1cm}   \textendash \textsc{LinUCB} & $0.292 \pm 0.113$ & $0.354 \pm 0.079$ & - \\

   \hspace{0.1cm}  \textendash \textsc{AlwaysQuery} & $0.727 \pm 0.195$ & $0.359 \pm 0.137$ & $1 \pm 0$ \\
\cline{1-4}
 \hspace{0.1cm}   \textendash \textsc{LinUCB-ExpDecay} & $0.303 \pm 0.116$ & $0.353 \pm 0.082$ & $0.160 \pm 0.150$ \\
 \hspace{0.1cm}   \textendash \textsc{LinUCB-QG} & $0.336 \pm 0.069$ & $0.342 \pm 0.069$ & $0.333 \pm 0.094$ \\
 \hline 
 \hline 
  Workload Model $f_{1,2,sim}$ &  &  &    \\
 \hspace{0.1cm}   \textendash \textsc{LinUCB} & $0.292 \pm 0.113$ & $0.343 \pm 0.079$ & - \\
   \hspace{0.1cm}  \textendash \textsc{AlwaysQuery} & $0.727 \pm 0.195$ & $0.359 \pm 0.137$ & $1 \pm 0$ \\
\cline{1-4}
 \hspace{0.1cm}   \textendash \textsc{LinUCB-ExpDecay} & $0.303 \pm 0.116$ & $0.330 \pm 0.097$ & $0.160 \pm 0.150$ \\
 \hspace{0.1cm}   \textendash \textsc{LinUCB-QG} & $0.503 \pm 0.041$ & $\mathbf{0.376 \pm 0.131}$ & $\mathbf{0.667 \pm 0.189}$ \\
 \hline
\end{tabular}
\vspace{1ex}
\vspace{-0.2cm}
\captionsetup{font=small}
\caption{Querying algorithms simulated on the test set, corresponding to $w_{task} = 0.7$, for 3 different data regimes in the workload model, averaged across 3 seeds. Values for $r_{task,avg}$ and $f_q$ correspond to the hyperparameter setting with maximal $M_{wt}$ on the validation set (for each separate data setting).}
\label{table:simulated-results}
\end{table}

In \textsc{LinUCB-QG}, the worst-case gap is the difference between a pessimistic estimate of $a^*$'s reward and an optimistic estimate of $a^{-}$'s reward, considering their confidence intervals (Fig. \ref{fig:hil-cb-pipeline}, top). A larger gap indicates a higher risk that the predicted best arm may be suboptimal, increasing the odds that the benefit of querying outweighs the workload penalty $WL_t$. Therefore, querying only when the gap is sufficiently large helps balance task reward and workload. 

We define the predicted workload penalty to be the counterfactual workload $WL_t$ if we were to query at the current time. To estimate this, we condition our workload model on the specific query type that we consider in our experiments. We use $WL_t = f(WL_0, Q_t; \theta)$, where we set the query type variables to be $d_t = \text{easy}$, $resp_t = \ac{MCQ}$, $dist_{t} = $ ``no distraction task" for all $t$ where the selected action $a_t = a_q$.

\vspace{-0.1cm}
\section{EXPERIMENTS}
\label{sec:experiments}

We evaluate our human-in-the-loop algorithm performance using two setups: (i) a simulation testbed (Sec. \ref{subsec:simulated-exps}), and (ii) a real world user study with 19 subjects (Sec. \ref{subsec:kinova-study}). We evaluate our algorithms on a surrogate objective, adapted from $\mathbb{E}[r_t|x_t, a_t, t]$ (Sec. \ref{sec:problem-formulation}) (details in Appendix\appendixcite{appendix}): 
\begin{align*}
    w_{task} \left[\frac{1}{T} \sum_{t=1}^T r_{task}(x_t,a_t)\right] - (1-w_{task}) (WL_T - WL_0)
\end{align*}


\vspace{-0.1cm}
\subsection{Simulated Testbed}
\label{subsec:simulated-exps}

\begin{figure*}[t]
    \centering
    \vspace{0.1cm}
    \includegraphics[width=0.95\textwidth]{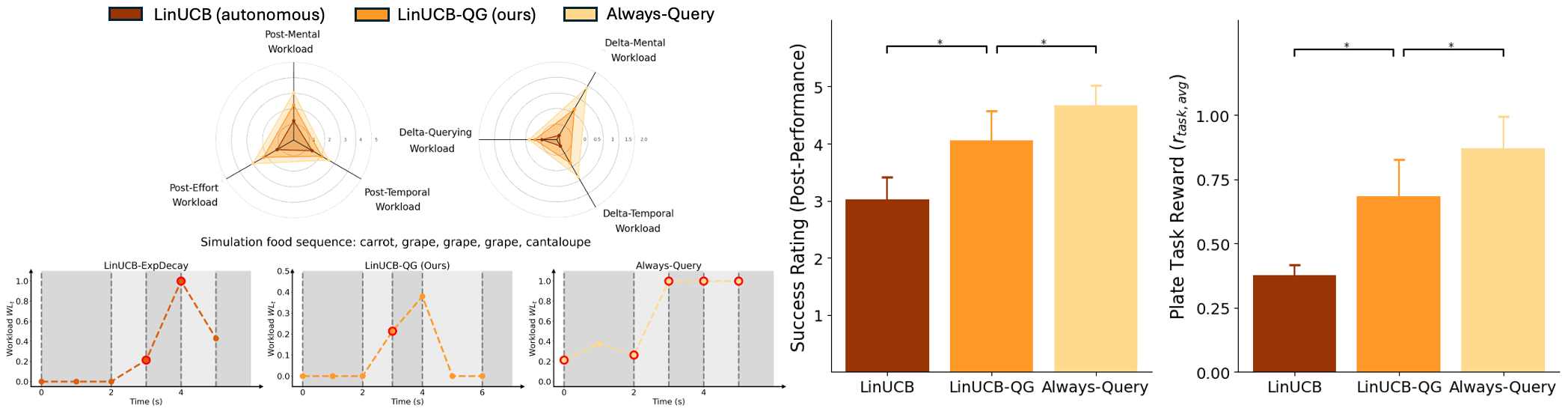}
    \captionsetup{font=small}
    \caption{Simulation and real-world results. \textit{Left (bottom):} Workload comparison between three querying algorithms in simulation ($D_2$ setting), with a sequence of 5 food items. Vertical lines indicate start of new food item, while red circles indicate query timesteps. \textsc{LinUCB-QG} reduces querying workload $WL_t$ compared to the other two methods, while offering competitive convergence times compared to \textsc{LinUCB-ExpDecay}. \textit{Left (top):} \textsc{LinUCB-QG} balances real-world performance and user workload, significantly reducing workload compared to \textsc{Always-Query} (which achieves slightly higher success at greater user cost), as shown by three post-method metrics (Post) and three change in workload metrics (Delta). \textit{Right:} \textsc{LinUCB-QG} significantly outperforms \textsc{LinUCB}, our autonomous baseline, in real-world bite acquisition success, both subjectively (Post-Performance) and objectively ($r_{task,avg}$). Error bars indicate standard error.}
    \label{fig:kinova-userstudyresult}
    \vspace{-0.4cm}
\end{figure*}

We simulate food interaction using a dataset of bite-sized food items on plates, including 16 food types \cite{feng2019robot}, each with 30 trials for the 6 robot actions $a_r$. We first draw a random image of a given food type from the dataset and generate the corresponding context $x$ using SPANet. When the policy selects a robot action $a_r$, we sample a binary task reward $r$ from a Bernoulli distribution with success probability $s_x(x,a_r)$. If $r=1$, we declare that the bandit has converged, and move to a new food type at the next timestep. If $r=0$, we draw a new random image of the same food type from the dataset until we exceed the maximum number of attempts $N_{att}=10$. When the policy selects a query action $a_q$, the bandit policy repeatedly executes the optimal action $a^*(x)$ until convergence, or the limit $N_{att}$ is reached.

\textbf{Metrics.} We compute 5 objective metrics to evaluate the tradeoff between task performance and querying workload:
\begin{itemize}[leftmargin=*,itemsep=0.0em,topsep=0pt]
    \item The mean task reward $r_{task, avg} = \frac{1}{T} \sum_{t=1}^{T} r_{task}(x_t,a_t)$, measuring the efficiency of successful food acquisition.
    \item The episodic change in workload $\Delta WL = WL_{T} - WL_0$.
    \item Our surrogate objective $M_{wt} = w_{task} r_{task, avg} - (1-w_{task}) \Delta WL$, where $w_{task}$ represents a user-specific preference for the task performance/workload tradeoff.
    \item The number of timesteps taken to converge $t_{conv}$, which provides additional insights into task performance.
    \item The fraction of food items for which we queried $f_q$, which provides an insight into querying workload.
\end{itemize}

\textbf{Evaluating querying workload in simulation.} In our simulation environment, we use the learned workload model $f$ for both $\textsc{LinUCB-QG}$ action-selection, as well as computing the evaluation metrics $\Delta WL$ and $M_{wt}$. During the rollout of a particular bandit policy, we evolve the workload model state $WL_t$ according to the learned model when we query. We consider the three dataset scenarios and corresponding workload models $f_{1,sim}$, $f_{2,sim}$, and $f_{1,2,sim}$ (Sec. \ref{subsec:predictive-model}), and set $WL_0 = 0.5$ (a median initial workload).

\textbf{Pretrain, validation, and test sets.} In our experiments, we partition the set of 16 food types into a pretraining set (for the contextual bandit and for training SPANet), a validation set (for tuning the querying algorithm hyperparameters), and a test set (for metric evaluation). Our validation set (cantaloupe, grape) and test set (banana, carrot) include food items with varying material properties. Our pretraining set includes the remaining 12 food types in the food dataset. We use the validation set to select the hyperparameter that maximizes the weighted metric $M_{wt}$ given a $w_{task}$ setting.

\textbf{Results.} We investigate how \textsc{LinUCB-QG} balances the tradeoff between task reward and querying workload. Table \ref{table:simulated-results} compares the four algorithms for a setting with a slight preference for maximizing task performance over minimizing workload ($w_{task} = 0.7$). Across all three workload data settings, \textsc{LinUCB-QG} has a higher mean task reward $r_{task,avg}$ than \textsc{LinUCB} and \textsc{LinUCB-ExpDecay}, and a mean weighted metric $M_{wt}$ that is better than or competitive with the other methods (within the uncertainty of the simulation), suggesting that it achieves the best overall tradeoff between task reward and querying workload. Additionally, the querying fraction $f_q$ for \textsc{LinUCB-QG} increases from $D_2$ to $D_{1,2}$ to $D_1$, suggesting that \textsc{LinUCB-QG} is sensitive to the higher workload predictions with care recipient and OT data, while still trading off task reward and workload. 

\textbf{Experiment: generalization on $D_2$.} We run an additional experiment to ablate $D_2$, where at test time we use either $f_{1,sim}$ or $f_{1,2,sim}$ in \textsc{LinUCB-QG} action-selection, but use $f_{2,sim}$ for metric evaluation ($w_{task}=0.4$). We find that $M_{wt}$ is lower in the first setting ($0.269\pm0.040$) compared to the second setting ($0.373\pm0.108$), suggesting that including $D_2$ is crucial to achieve the best trade-off between task reward and workload for users with mobility limitations. See Appendix\appendixcite{appendix} for more details and comparison metrics.

\vspace{-0.1cm}
\subsection{Real-World User Study}
\label{subsec:kinova-study}
\vspace{-0.1cm}

To evaluate the real-world performance-workload tradeoff of \textsc{LinUCB-QG}, we conduct a user study with 19 users: 18 users without mobility limitations (8 male, 10 female; ages 19-31; 66\% with prior robot interaction experience), and one 46-year old user with Multiple Sclerosis since they were 19. We investigate whether \textsc{LinUCB-QG} improves task performance compared to \textsc{LinUCB}, while minimizing workload compared to \textsc{Always-Query}. This study was approved by the IRB at Cornell University (Protocol \#IRB0010477). 

\textbf{Study setup.} We use a Kinova Gen-3 6-DoF robot arm with a custom feeding tool (details in Appendix\appendixcite{appendix}). For each method, we present users with a plate of 3 food items (banana, carrot, cantaloupe), with diverse characteristics: cantaloupe works with all robot actions; carrots require sufficient penetration force and fork tines perpendicular to the food major axis; bananas are soft, requiring tilted-angled actions \cite{gordon2020adaptive}. We set $N_{att}=3$ for a reasonable study duration. The user provides feedback using a speech-to-text interface when queried, without any additional distraction tasks. Users evaluate each of the three methods in a counterbalanced order. We measure workload before/after each method by asking 5 modified NASA-TLX subscale questions: mental/physical, temporal, performance, effort, and frustration (question wording in Appendix\appendixcite{appendix}). We conduct 2 repetitions of the 3 food items for all 3 methods (18 total trials). 

\textbf{Metrics.} We define 7 subjective and 3 objective metrics to compare the methods. 4 subjective metrics correspond to the modified post-method NASA-TLX questions: Post-Mental Workload, Post-Temporal Workload, Post-Performance, and Post-Effort Workload. The other 3 subjective metrics measure workload changes during each method: Delta-Mental Workload, Delta-Temporal Workload, and Delta-Querying Workload (the change in querying workload score, weighting function in Appendix\appendixcite{appendix}). The 3 objective metrics are mean task reward per plate ($r_{task,avg}$), mean successes per plate ($n_{success}$), and mean query timesteps per plate ($n_{q}$).

\textbf{Overall results.} Among the three methods, \textsc{LinUCB-QG} offers the most balanced approach to human-in-the-loop bite acquisition. It is more efficient than \textsc{LinUCB} and imposes less querying workload than \textsc{Always-Query}. While \textsc{LinUCB} typically picks up the carrot or cantaloupe within 1-2 timesteps but struggles with the banana, \textsc{Always-Query} successfully picks up food in the first timestep by always querying the user, leading to higher workload. Our method, \textsc{LinUCB-QG}, selectively asks for help with the banana and acts autonomously for the other foods in most trials, balancing task performance with querying workload.

\textit{(1) Task success results:} Fig. \ref{fig:kinova-userstudyresult} (right) shows that \textsc{LinUCB-QG} achieves higher objective ($r_{task,avg}$) and subjective (Post-Performance) success ratings versus \textsc{LinUCB}, our autonomous baseline, showing greater efficiency. \textit{(2) Querying workload results:} As shown in Fig. \ref{fig:kinova-userstudyresult} (left, top), \textsc{LinUCB-QG} has lower subjective querying workload scores (mental, temporal, effort) compared to \textsc{Always-Query}, indicating that selective querying reduces workload. This is the case for both the post-method metrics (Post) and the workload change metrics (Delta). All results are statistically significant (Wilcoxon paired signed-rank test, $\alpha = 0.05$). See Appendix for full comparisons across all metrics\appendixcite{appendix}.

\textbf{Results: user with mobility limitations.} We highlight observations from the user with mobility limitations. First, they provided a higher mean subjective success rating for \textsc{LinUCB-QG} than for \textsc{LinUCB}, aligning with the aggregate results. However, the user provided consistently low physical/mental, temporal and effort ratings across all methods. Potential reasons for this include the relative simplicity of providing feedback in our setting, and the fact that this particular user frequently requests assistance in daily life, reducing their querying workload. The user also commented that relying on external physiological signals (such as EEG) to estimate workload would add stress (due to the additional hardware required), reinforcing the benefit of non-intrusive workload models for human-in-the-loop querying algorithms.

\textbf{Discussion.} Future work could incorporate different query types in the real-world study, beyond the best acquisition action, such as the other query types in our workload model. In our setting, task complexity depends on physical food properties such as compliance or shape. We did not explicitly focus on such variables when modeling workload, although query difficulty $d_i$ does correlate with task complexity. The NASA-TLX scale has known limitations, such as subjectivity \cite{hart2006nasa}, over-emphasizing task difficulty \cite{mckendrick2018deeper}, and workload score calculation problems [\citeconsecutive{virtanen2022weight, bolton2023mathematical}], meaning that the modified NASA-TLX scores used to train our workload models are inherently noisy. We mitigate this by focusing on relative workload changes when evaluating how well our algorithms balance performance and workload. Additionally, we do not take measurements between queries in our workload dataset, so our workload models interpolate between queries using their learned weights, leading to rollouts that are not always intuitive (e.g. due to assigning greater weight to past queries). Future directions include predictive modeling with objective, non-intrusive workload measurements, and extensions to other assistive tasks where performance must be balanced with workload, using domain-agnostic workload features.

\newpage
\printbibliography


\section*{APPENDIX}
\label{sec:appendix}

\subsection{Problem Formulation}

\subsubsection{Details on Contextual Bandit Setting}

Below we provide more details on the contextual bandit setting.

\begin{itemize}[leftmargin=*,itemsep=-0.1em]
    \item \textbf{Observation space $\mathcal{O}$}: In each RGB image, the food item is either isolated, close to the plate edge, or on top of another food item \cite{feng2019robot}.
    \item \textbf{Action space $\mathcal{A}$}: Each robot action $a_r$ is a pair consisting of one of three pitch configurations (tilted angled (TA), vertical skewer (VS), tilted vertical (TV)) and one of two roll configurations ($0^\circ$, $90^\circ$), relative to the orientation of the food [\citeconsecutive{bhattacharjee2019towards, feng2019robot}].
    \item \textbf{Context space $\mathcal{X}$}: SPANet, the network from which we derive our contexts $x$, is pretrained in a fully-supervised manner to predict $s_o(o,a)$, the probability that action $a$ succeeds for observation $o$. 
\end{itemize}

\subsection{Querying Workload User Study} 

\subsubsection{NASA-TLX weighting function}

Throughout the paper, we define the querying workload state $WL$ to be the following weighted sum of the raw NASA-TLX subscales: $WL = 0.4 \cdot \text{mental} + 0.2 \cdot \text{temporal} + 0.4 \cdot \text{effort}$.

\subsubsection{Modified NASA-TLX Subscale Questions}

Below are the question wordings that we use in our online workload user study, which are modified versions of the NASA-TLX subscales:

\begin{itemize}
    \item Mental/Physical Demand: ``How mentally or physically demanding was the task?"
    \item Temporal Demand: ``How hurried or rushed was the pace of the task?"
    \item Performance: ``How successful were you in accomplishing what you were asked to do?"
    \item Effort: ``How hard did you have to work to accomplish your level of performance?"
    \item Frustration: ``How irritated, stressed, and annoyed were you?"
\end{itemize}
    
Note that for the users without mobility limitations, the wording of the Mental/Physical Demand task was the following: ``How mentally demanding was the task?". This is because we assumed that the physical demand to answer the workload questions was minimal for users without mobility limitations.

\subsubsection{Baseline conditions} 2 baseline conditions. Each baseline condition corresponds to a fixed setting of $dist_t$ (either easy addition, or hard video transcription), without any robot query tasks. 

\subsubsection{Sample study questions} Below we provide examples of the robot queries that we ask in the online querying workload user study, covering each of the possible response types $r_t$ and question difficulties $d_t$:

\begin{itemize}[leftmargin=*,itemsep=-0.1em]
    \item $r_t = \text{MCQ}$, $d_t = \text{easy}$:  ``What kind of food item is outlined in the image below?" (Responses: ``Raspberry", ``Strawberry", ``Grape", ``Apple", ``I don't  know.")
    \item $r_t = \text{BB}$, $d_t = \text{easy}$: ``Draw a box around only the strawberry."
    \item $r_t = \text{OE}$, $d_t = \text{easy}$: ``Why did the Robot fail in acquiring the cantaloupe?"
    \item $r_t = \text{MCQ}$, $d_t = \text{hard}$: ``Which of the following images is tofu?" (Responses: ``Left", ``Right", ``I don't know")
    \item $r_t = \text{BB}$, $d_t = \text{hard}$: ``Draw a box around only the carrot in the bottom right."
    \item $r_t = \text{OE}$, $d_t = \text{hard}$: ``How would you skewer the following item with a fork?"
\end{itemize}

\subsubsection{Compensation} Users without mobility limitations were compensated at a rate of \$12/hr, for a total compensation of \$18 for the entire study (as the study duration was 1.5 hr). The occupational therapists who simulated mobility limitations were compensated at a rate of \$10/hr, for a total compensation of \$15 for the entire study. The users with mobility limitations were compensated at a rate of \$20/hr, for a total compensation of \$30 for the entire study.

\subsubsection{Mean querying workload difference, $D_1$ vs $D_2$.} In Section \ref{subsec:workload-data-analysis}, we describe finding a statistically significant difference in mean querying workload between $D_1$ (users without limitations) and $D_2$ (users with limitations and OTs), where $D_1$ had a mean querying workload of $0.36 \pm 0.29$, and $D_2$ had a mean querying workload of $0.40 \pm 0.28$. We run a Mann-Whitney U-test on the querying workload scores from $D_1$ and $D_2$. The alternative hypothesis that we tested is that the distribution of workload values for $D_1$ is stochastically less than the distribution of workload values for $D_2$. For this alternative hypothesis, we find that $p = 5.410\text{e-}05$, indicating a statistically significant result for $\alpha = 0.05$.

\subsection{Querying Workload Predictive Model}

\subsubsection{Modified TLX survey timing information}

When training the workload models, we make the following assumption about the amount of time taken for users to complete the modified TLX surveys. In the $D_1$ setting, we assume 0 seconds for the full user population (which only includes users without mobility limitations). In the $D_{1,2}$ setting, we assume 5 seconds for users without mobility limitations (and for two users in the $D_2$ population for whom we did not collect time data), and for the rest of the users with mobility limitations, we used the logged time taken to complete the surveys. Our justification for this design choice is that in the $D_{1,2}$ setting, we would like the assumed time for users without mobility limitations to be appropriately scaled compared to the logged times for users with limitations, who took non-zero times to complete the surveys.

\subsubsection{Discrete-time Models} We consider different values of $H$, with one memoryless Granger model (where $H=1$) and a set of Granger models with $H$ ranging from $5$ to $30$. We chose the upper bound of $H=30$ to roughly correspond to the study condition length in Section \ref{subsec:user-study}. This is because we set each discrete time step to correspond to $10\text{s}$. We zero-pad the query variables $[d_{t-i}, resp_{t-i}, dist_{t-i}]$ if there are no queries in the discrete time window $t-i$, or if we run out of history.

For each of these models, we also consider variants where we impose non-negativity constraints and/or ridge-regression penalties on the weights $\gamma$ and $\mathbf{w_i}^T$. Specifically, we consider 4 different variants: 

\begin{itemize}
    \item Granger: no nonnegativity or ridge-regression penalty
    \item Granger-Nonnegative (Granger-N): nonnegativity constraint only
    \item Granger-Ridge (Granger-R): ridge-regression penalty only
    \item Granger-Ridge-Nonnegative (Granger-RN): non-negativity constraint and ridge-regression penalty
\end{itemize}

\subsubsection{Continuous-time Models} We also consider a continuous-time setting, where $WL(t)$ represents the querying workload at time $t$. In this setting, we consider one model (denoted as Exp-Impulse) that models the workload at the current time $t$ as composed of a series of impulses at the query times, with an exponential decay in workload in between the queries. The Exp-Impulse model is defined recursively as follows:
\begin{equation*}
    WL(t) = WL(t_{prev}) e^{-\lambda (t - t_{prev})} + \beta(d_t, r_t, dist_t)
\end{equation*}
where $t_{prev}$ is the previous timestep at which we queried, $\beta(d_t, r_t, dist_t)$ is the magnitude of the impulse, and $\lambda$ is the workload decay rate.

\subsubsection{Model Selection} 
For each of the three dataset settings ($D_1$, $D_2$, $D_{1,2}$), we perform 4-fold cross-validation to select a model from the above set of linear models. Dataset $D_1$ consists of $N=4272$ unique question/workload pairs, while dataset $D_2$ consists of $N=705$ unique question/workload pairs. For tuning of the ridge regression regularization parameter $\lambda_r$, we perform 5-fold cross validation to select the optimal parameter value, with a logarithmic range from $10^{-3}$ to $10^2$.

Table \ref{table:cogload-models} shows predictive MSE statistics on the held-out test splits, computed across the 4 folds, for the set of learned models. Note that we also considered a constant baseline (denoted Constant), whose predicted workload is $WL_t = WL_0$, and an average baseline (denoted Average), whose predicted workload is the average value of $WL_0$ in the training set. 

\textit{Real-world user study models.} We describe the two simulated workload models ($f_{1,real}$, $f_{1,2,real}$) that are used for the real-world experiments in Section \ref{subsec:kinova-study}, along with the mobility limitation-only model $f_{2,real}$ that is described in Section \ref{subsec:predictive-model}. The model $f_{1,real}$ is a Granger model with $H=5$, with a ridge-regression penalty on $\gamma$ and $\mathbf{w_i}^T$, where the final selected ridge regression penalty was $\lambda_r = 10$. The model $f_{2,real}$ is a Granger model with $H=1$. The model $f_{1,2,real}$ is a Granger model with $H=1$ and a ridge-regression penalty on $\gamma$ and $\mathbf{w_i}^T$, where the final selected ridge regression penalty was $\lambda_r = 10$. The model $f_{1,real}$ achieves a cross-validation mean test MSE of $0.0573 \pm 0.0127$ on $D_1$, while $f_{1,2,real}$ achieves a cross-validation mean test MSE of $0.0582 \pm 0.0099$ on the $D_{1,2}$ dataset.

We use these models because they are the models with the lowest median test MSE on each dataset. We initially considered using mean test MSE to select the best-performing model, but we discovered that the mean test MSE had a very large magnitude for certain models. This is because for the model settings that do not have weight constraints, linear regression would overfit to the training set and learn model parameters with large weight magnitudes. Because of this, we use the median test MSE to select the model. Note that while the models have differences in their mean test MSE, the standard deviations in test MSE for the models overlap because of the variance across folds.

\textit{Simulated models.} We describe the three simulated workload models ($f_{1,real}$, $f_{2,real}$, $f_{1,2,real}$) that are used in the experiments in Section \ref{subsec:simulated-exps}. For all three models, we use Granger models with $H=10$, where we place the following constraints on the parameters during model training: $\gamma, \{\mathbf{w_i}^T\}_{i=0}^{H-1} \in [0.05, 1]$, $w_0 \leq 1$. Placing these additional constraints guarantees that the predicted change in workload associated with a query is non-negative.  

\begin{table*}[h]
\centering
\begin{tabular}{ |p{0.75in}||c|c|c|c|c|c|c| } 
\hline
 & & \multicolumn{2}{c}{$D_1$} & \multicolumn{2}{|c}{$D_2$} & \multicolumn{2}{|c|}{$D_{1,2}$} \\
\hline
\centering \textbf{Model} & $H$ & Test MSE ($\mu \pm \sigma$) & \makecell{Test MSE \\ (median)} & Test MSE ($\mu \pm \sigma$) & \makecell{Test MSE \\ (median)} & Test MSE ($\mu \pm \sigma$) & \makecell{Test MSE \\ (median)} \\ 
 \hline
 \hline
Constant & 0 & $0.1271 \pm  0.0173$ & $0.1304$ & $0.104 \pm 0.0245$ & $0.112$ & $0.123 \pm 0.0164$ & $0.124$ \\
Average & 0 & $0.0937 \pm  0.0086$ & $0.0922$ & $0.089 \pm 0.0319$ & $0.0933$& $0.0919 \pm 0.00806$ & $0.0944$ \\
\hline
\hline
\multirow{7}{*}{Granger} & 1 & $0.0572 \pm  0.0126$ & $0.0507$ & $0.066 \pm 0.0219$ & $\mathbf{0.0761}$ & $0.0583 \pm 0.00988$ & $0.0573$ \\
& 5 & $7.59\text{\text{e+}}15 \pm  1.31\text{\text{e+}}16$ & $0.0656$ & $3.95\text{e+}23 \pm 6.68\text{e+}23$ & $1.38\text{e+}22$& $3.66\text{e+}16 \pm 6.34\text{e+}16$ & $0.0581$ \\
&10 & $3.16\text{e+}18 \pm  5.47\text{e+}18$ & $0.0661$ & $1.08\text{e+}24 \pm 1.88\text{e+}24$ & $2.24\text{e+}20$& $3.38\text{e+}15 \pm 5.86\text{e+}15$ & $0.058$ \\
&15 & $1.87\text{e+}17 \pm  3.23\text{e+}17$ & $0.0660$ & $2.08\text{e+}21 \pm 3.23\text{e+}21$ & $3.23\text{e+}20$& $3.45\text{e+}15 \pm 5.98\text{e+}15$ & $0.0582$ \\
&20 & $5.41\text{e+}17 \pm  9.37\text{e+}17$ & $0.0663$ & $2.43\text{e+}24 \pm 4.21\text{e+}24$ & $1.72\text{e+}21$& $1.92\text{e+}12 \pm 3.32\text{e+}12$ & $0.0584$ \\
&25 & $1.39\text{e+}18 \pm  2.41\text{e+}18$ & $0.0668$ & $5.05\text{e+}23 \pm 7.95\text{e+}23$ & $7.11\text{e+}22$& $3.07\text{e+}21 \pm 5.32\text{e+}21$ & $0.0587$ \\
&30 & $1.09\text{e+}20 \pm  1.89\text{e+}20$ & $0.0677$ & $1\text{e+}24 \pm 1.34\text{e+}24$ & $3.64\text{e+}23$ & $5.78\text{e+}21 \pm 1\text{e+}22$ & $0.059$ \\
\hline
\multirow{7}{*}{Granger-N} & 1 &  $0.0573 \pm  0.0126$ & $0.0507$ & $0.0662 \pm 0.0217$ & $0.0762$& $0.0584 \pm 0.00997$ & $0.0576$ \\
& 5 & $0.0575 \pm  0.0128$ & $0.0508$ & $0.0665 \pm 0.0215$ & $0.0763$ & $0.0586 \pm 0.0101$ & $0.0576$ \\
& 10 & $0.0575 \pm  0.0128$ & $0.0508$ & $0.0682 \pm 0.0199$ & $0.0769$& $0.0585 \pm 0.00997$ & $0.0576$ \\
& 15 & $0.0574 \pm  0.0126$ & $0.0508$ & $0.0698 \pm 0.0196$ & $0.0784$ & $0.0586 \pm 0.00985$ & $0.0578$ \\
& 20 & $0.0575 \pm  0.0127$ & $0.0508$ & $0.0705 \pm 0.0204$ & $0.0798$& $0.0588 \pm 0.00982$ & $0.0581$ \\
& 25 & $0.0577 \pm  0.0126$ & $0.0508$ & $0.0726 \pm 0.0211$ & $0.0819$& $0.0589 \pm 0.00991$ & $0.0582$ \\
& 30 & $0.0578 \pm  0.0125$ & $0.0510$ & $0.0735 \pm 0.0205$ & $0.083$ & $0.0591 \pm 0.00994$ & $0.0583$ \\
\hline
\multirow{7}{*}{Granger-R} & 1 & $0.0572 \pm  0.0127$ & $0.0507$ & $0.066 \pm 0.0219$ & $0.0763$& $0.0582 \pm 0.00988$ & $\mathbf{0.0572}$ \\
& 5 & $0.0573 \pm  0.0127$ & $\mathbf{0.0506}$ & $0.0663 \pm 0.0216$ & $0.0772$ & $0.0584 \pm 0.0099$ & $0.0576$ \\
& 10 & $0.0574 \pm  0.0125$ & $0.0509$ & $0.0675 \pm 0.0217$ & $0.0783$ & $0.0584 \pm 0.00964$ & $0.0574$ \\
& 15 & $0.0574 \pm  0.0124$ & $0.0512$ & $0.0691 \pm 0.0219$ & $0.0801$& $0.0584 \pm 0.00948$ & $0.0575$ \\
& 20 & $0.0575 \pm  0.0124$ & $0.0512$ & $0.0687 \pm 0.0221$ & $0.0795$ & $0.0588 \pm 0.00939$ & $0.0582$ \\
& 25 & $0.0579 \pm  0.0123$ & $0.0514$ & $0.0694 \pm 0.0223$ & $0.0801$ & $0.0589 \pm 0.00951$ & $0.0583$ \\
& 30 & $0.0582 \pm  0.0122$ & $0.0515$ & $0.0694 \pm 0.0222$ & $0.0801$& $0.0591 \pm 0.00956$ & $0.0583$ \\
\hline
\multirow{7}{*}{Granger-RN} & 1 & $0.0573 \pm  0.0126$ & $0.0508$ & $0.0719 \pm 0.0246$ & $0.0819$& $0.0589 \pm 0.0105$ & $0.0574$ \\
& 5 & $5.66\text{e+}24 \pm  9.80\text{e+}24$ & $0.0655$ & $9.96\text{e+}21 \pm 1.73\text{e+}22$ & $1.6\text{e+}15$ & $5.88\text{e+}24 \pm 1.02\text{e+}25$ & $4.02\text{e+}15$ \\
& 10 & $0.6120 \pm  0.9559$ & $0.0658$ & $5.16\text{e+}14 \pm 8.94\text{e+}14$ & $0.0894$& $0.0607 \pm 0.0112$ & $0.061$ \\
& 15 & $0.0573 \pm  0.0125$ & $0.0508$ & $5.62\text{e+}12 \pm 9.73\text{e+}12$ & $0.0862$ & $0.0589 \pm 0.0104$ & $0.0574$ \\
& 20 & $0.0574 \pm  0.0125$ & $0.0509$ & $3.46\text{e+}13 \pm 3.49\text{e+}13$ & $3.15\text{e+}13$ & $4.45\text{e+}04 \pm 7.71\text{e+}04$ & $0.0676$ \\
& 25 & $0.1189 \pm  0.1179$ & $0.0514$ & $1.1\text{e+}14 \pm 1.9\text{e+}14$ & $0.0895$ & $0.059 \pm 0.0104$ & $0.0575$ \\
& 30 & $0.1195 \pm  0.1021$ & $0.0662$ & $1.92\text{e+}14 \pm 2.7\text{e+}14$ & $5.65\text{e+}13$ & $0.059 \pm 0.0104$ & $0.0575$\\
 \hline
 \hline
Exp-Impulse & - &  $0.0737 \pm 0.0128$ & $0.0678$ & $0.0809 \pm 0.0199$& $0.0878$ & $0.0753 \pm 0.0112$ & $0.0766$\\
\hline
\end{tabular}
\caption{Test mean-squared error (MSE) values for querying workload models, for all three workload data settings ($D_1$, $D_2$, $D_{1,2})$. For each data setting, the model with the lowest median test MSE is indicated in bold.}
\label{table:cogload-models}
\end{table*}

\subsection{Human-in-the-Loop Algorithms} We provide the full algorithmic description for \textsc{LinUCB-ExpDecay} in Algorithm \ref{alg:exp-decay-query}.

\begin{algorithm}
\caption{\textsc{LinUCB-ExpDecay}.}
\label{alg:exp-decay-query}
\begin{algorithmic}[1]
\State Inputs: Context $x$, decay rate $c$, number of food items seen $N$, time $t$
\State For all robot actions $a_r$, compute UCB bonus $b_{a_r}$ and UCB value $UCB_{a_r}$.
\State{Set $P(query) = e^{-cN}$ if $t$ is the first timestep to observe $x$,  $P(query) = 0$ otherwise.}
\State{Set $a = a_q$ with probability $P(query)$, $a = \arg \max_{a_r} UCB_{a_r}$ with probability $1-P(query)$.} \\
\Return $a$
\end{algorithmic}
\end{algorithm}

\begin{figure}[t]
  \centering
    \includegraphics[scale=0.35]{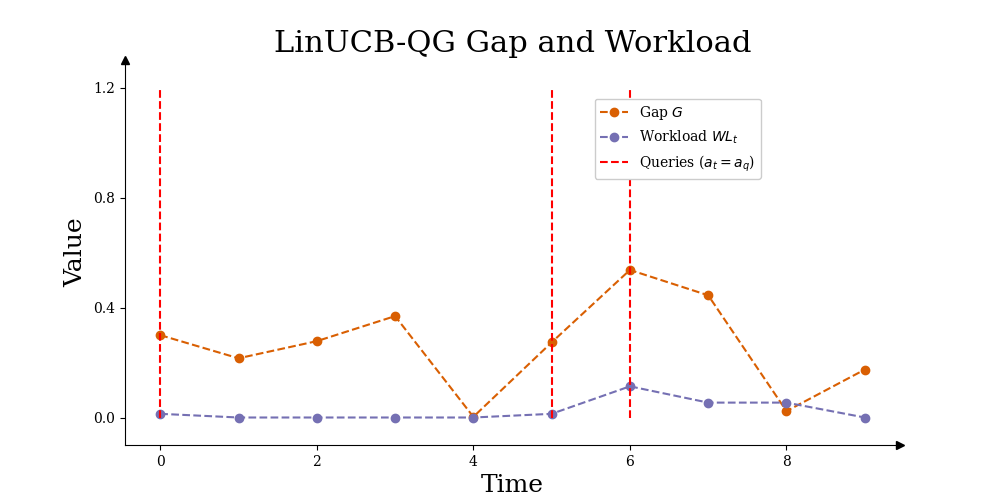}
    \caption{\small Example LinUCB-QG rollout in simulation environment, illustrating evolution of performance gap $G$ and workload $WL_t$, along with timesteps corresponding to query times. LinUCB-QG decides to ask the human for help when $G > w WL_t$ (in this rollout, $w = 4$).}
    \label{fig:appendix-gap}
\end{figure}

\subsection{Experiments: Simulated Testbed} 

\subsubsection{Surrogate Objective}

Here we justify our choice of the surrogate objective outlined in Section \ref{sec:experiments}. Recall that the expected reward $\mathbb{E}[r_t|x_t, a_t, t]$ defined in Section \ref{sec:problem-formulation} depends explicitly on the querying workload $WL_t$ for all times $t$ for which $a_t = a_q$. However, measuring the intermediate workload values after every query would require administering a survey after each query, which is impractical in the real world. Therefore, in our experimental formulation in both simulation and in the real study, we will assume that we cannot observe the intermediate workload values. Instead, we will observe only the initial and final workload workload values, which are $WL_0$ and $WL_T = f(WL_0, Q_T; \theta)$, respectively.

\subsubsection{Additional Results}

First, we include a set of additional metrics for the experimental setting described in Section \ref{subsec:simulated-exps}, which focus on the observed convergence for each food item. We define the following metrics: $f_{fail,food}$, which is the fraction of food items for which the algorithm was unable to converge; and $f_{auto,food}$, which is the fraction of food items for which the algorithm autonomously converged. We also show $\Delta WL$, the change in querying workload across an episode and $t_{conv}$, the number of timesteps required to converge to the optimal action. Table \ref{table:addnl-results-main-body} shows these metrics for $w_{task}=0.4$.

\begin{table*}[t]
\vspace{0.2cm}
\centering
\begin{tabular}{ |p{1.25 in}||c|c|c|c| } 
\hline
\textbf{} & $\Delta WL$ & $t_{conv}$ & $f_{fail,food}$ & $f_{auto,food}$ \\ 
 \hline
 \hline
Workload Model $f_{1,sim}$ &  &   & &\\
\hspace{0.1cm}   \textendash  \textsc{LinUCB} & $-0.500 \pm 0.000$ & $3.467 \pm 1.087$ & $0.067 \pm 0.094$ & $0.933 \pm 0.094$ \\
\hspace{0.1cm}   \textendash \textsc{AlwaysQuery} & $-0.274 \pm 0.024$ & $1.467 \pm 0.340$ & $0.000 \pm 0.000$ & $0.000 \pm 0.000$ \\
   \cline{1-5}
\hspace{0.1cm}   \textendash \textsc{LinUCB-ExpDecay} & $-0.485 \pm 0.037$ & $2.413 \pm 0.463$ & $0.040 \pm 0.080$ & $0.400 \pm 0.146$ \\
\hspace{0.1cm}   \textendash \textsc{LinUCB-QG} & $-0.414 \pm 0.122$ & $1.667 \pm 0.499$ & $0.000 \pm 0.000$ & $0.133 \pm 0.189$ \\
 \hline
 \hline
 Workload Model $f_{2,sim}$ &  &  &    &\\
\hspace{0.1cm}   \textendash \textsc{LinUCB} & $-0.500 \pm 0.000$ & $3.467 \pm 1.087$ & $0.067 \pm 0.094$ & $0.933 \pm 0.094$ \\
\hspace{0.1cm}   \textendash \textsc{AlwaysQuery} & $0.500 \pm 0.000$ & $1.467 \pm 0.340$ & $0.000 \pm 0.000$ & $0.000 \pm 0.000$ \\
   \cline{1-5}
\hspace{0.1cm}   \textendash \textsc{LinUCB-ExpDecay} & $-0.471 \pm 0.107$ & $3.297 \pm 1.048$ & $0.067 \pm 0.094$ & $0.773 \pm 0.177$ \\
\hspace{0.1cm}   \textendash \textsc{LinUCB-QG} & $-0.357 \pm 0.101$ & $3.133 \pm 0.754$ & $0.000 \pm 0.000$ & $0.667 \pm 0.094$ \\
 \hline 
 \hline 
  Workload Model $f_{1,2,sim}$ &  &  &    &\\
 \hspace{0.1cm}   \textendash \textsc{LinUCB} & $-0.463 \pm 0.000$ & $3.467 \pm 1.087$ & $0.067 \pm 0.094$ & $0.933 \pm 0.094$ \\
\hspace{0.1cm}   \textendash \textsc{AlwaysQuery} & $0.500 \pm 0.000$ & $1.467 \pm 0.340$ & $0.000 \pm 0.000$ & $0.000 \pm 0.000$ \\
   \cline{1-5}
\hspace{0.1cm}   \textendash \textsc{LinUCB-ExpDecay} & $-0.395 \pm 0.240$ & $3.297 \pm 1.048$ & $0.067 \pm 0.094$ & $0.773 \pm 0.177$ \\
\hspace{0.1cm}   \textendash \textsc{LinUCB-QG} & $-0.078 \pm 0.411$ & $2.000 \pm 0.163$ & $0.000 \pm 0.000$ & $0.333 \pm 0.189$ \\
 \hline
\end{tabular}
\vspace{1ex}
\caption{Additional convergence metrics in simulation on the test set, corresponding to $w_{task} = 0.7$, for 3 different data regimes in the workload model, averaged across 3 seeds (For \textsc{LinUCB-ExpDecay}, we also average across 5 policy random seeds). Metric values for \textsc{LinUCB-ExpDecay} and \textsc{LinUCB-QG} correspond to the hyperparameter setting with maximal $M_{wt}$ on the validation set (for each separate data setting).}
\label{table:addnl-results-main-body}
\end{table*}

Next, we include full results for multiple settings of $w_{task}$ and $w_{cl}$, for the same querying workload models used in Section \ref{subsec:simulated-exps}, ranging from $w_{task} \in \{0.2, 0.3, \dots, 0.9\}$ shown in Tables \ref{table:wtask-0.2-to-0.9-f1}, \ref{table:wtask-0.2-to-0.9-f2}, and \ref{table:wtask-0.2-to-0.9-f12} (excluding $w_{task}=0.4$, whose results are shown in Table \ref{table:simulated-results}). In the $f_{1,sim}$ setting, we see that \textsc{LinUCB-QG} generally performs the best for intermediate values of $w_{task}$ (corresponding to an intermediate emphasis on minimizing workload), while \textsc{Always-Query} performs the best for higher values of $w_{task}$ (corresponding to a high emphasis on maximum task performance, which the \textsc{Always-Query} baseline achieves due to its lack of exploration compared to \textsc{LinUCB-QG}). In the $f_{2,sim}$ and $f_{1,2,sim}$ settings, \textsc{LinUCB} performs the best for lower values of $w_{task}$ (because it never asks for help), while \textsc{Always-Query} performs the best for higher values of $w_{task}$ (where task performance is more critical). Nevertheless, in all data settings and across different $w_{task}$ values, \textsc{LinUCB-QG} offers better mean $r_{task}$ compared to \textsc{LinUCB} and \textsc{LinUCB-ExpDecay}, and competitive $M_{wt}$ compared to the other methods.

\begin{table*}[t]
\centering
\begin{tabular}{ |c|p{1.125 in}||c|c|c||c|c| } 
\hline
\centering $w_{task}$ & \textbf{Method} & $r_{task,avg}$ & $M_{wt}$ & $f_{q}$ \\ 
 \hline
 \hline
\multirow{4}{*}{0.2}&\textsc{LinUCB} & $0.292 \pm 0.113$ & $0.458 \pm 0.023$ & $0.000 \pm 0.000$ \\
&\textsc{AlwaysQuery} & $0.727 \pm 0.195$ & $0.365 \pm 0.021$ & $1.000 \pm 0.000$ \\
\cline{2-5}&\textsc{LinUCB-ExpDecay} & $0.374 \pm 0.143$ & $\mathbf{0.475 \pm 0.029}$ & $0.440 \pm 0.150$ \\
&\textsc{LinUCB-QG} & $0.434 \pm 0.049$ & $0.458 \pm 0.049$ & $0.733 \pm 0.094$ \\

\hline
\hline
\multirow{4}{*}{0.3}&\textsc{LinUCB} & $0.292 \pm 0.113$ & $0.438 \pm 0.034$ & $0.000 \pm 0.000$ \\
&\textsc{AlwaysQuery} & $0.727 \pm 0.195$ & $0.410 \pm 0.042$ & $1.000 \pm 0.000$ \\
\cline{2-5}&\textsc{LinUCB-ExpDecay} & $0.374 \pm 0.143$ & $\mathbf{0.462 \pm 0.043}$ & $0.440 \pm 0.150$ \\
&\textsc{LinUCB-QG} & $0.434 \pm 0.049$ & $0.455 \pm 0.048$ & $0.733 \pm 0.094$ \\

\hline
\hline
\multirow{4}{*}{0.4}&\textsc{LinUCB} & $0.292 \pm 0.113$ & $0.417 \pm 0.045$ & $0.000 \pm 0.000$ \\
&\textsc{AlwaysQuery} & $0.727 \pm 0.195$ & $\mathbf{0.455 \pm 0.064}$ & $1.000 \pm 0.000$ \\
\cline{2-5}&\textsc{LinUCB-ExpDecay} & $0.374 \pm 0.143$ & $0.449 \pm 0.057$ & $0.440 \pm 0.150$ \\
&\textsc{LinUCB-QG} & $0.434 \pm 0.049$ & $0.452 \pm 0.047$ & $0.733 \pm 0.094$ \\

\hline
\hline
\multirow{4}{*}{0.5}&\textsc{LinUCB} & $0.292 \pm 0.113$ & $0.396 \pm 0.056$ & $0.000 \pm 0.000$ \\
&\textsc{AlwaysQuery} & $0.727 \pm 0.195$ & $0.501 \pm 0.086$ & $1.000 \pm 0.000$ \\
\cline{2-5}&\textsc{LinUCB-ExpDecay} & $0.374 \pm 0.143$ & $0.437 \pm 0.071$ & $0.440 \pm 0.150$ \\
&\textsc{LinUCB-QG} & $0.670 \pm 0.237$ & $\mathbf{0.542 \pm 0.059}$ & $0.867 \pm 0.189$ \\

\hline
\hline
\multirow{4}{*}{0.6}&\textsc{LinUCB} & $0.292 \pm 0.113$ & $0.375 \pm 0.068$ & $0.000 \pm 0.000$ \\
&\textsc{AlwaysQuery} & $0.727 \pm 0.195$ & $0.546 \pm 0.108$ & $1.000 \pm 0.000$ \\
\cline{2-5}&\textsc{LinUCB-ExpDecay} & $0.374 \pm 0.143$ & $0.424 \pm 0.086$ & $0.440 \pm 0.150$ \\
&\textsc{LinUCB-QG} & $0.670 \pm 0.237$ & $\mathbf{0.567 \pm 0.094}$ & $0.867 \pm 0.189$ \\

\hline
\hline
\multirow{4}{*}{0.8}&\textsc{LinUCB} & $0.292 \pm 0.113$ & $0.333 \pm 0.090$ & $0.000 \pm 0.000$ \\
&\textsc{AlwaysQuery} & $0.727 \pm 0.195$ & $\mathbf{0.636 \pm 0.152}$ & $1.000 \pm 0.000$ \\
\cline{2-5}&\textsc{LinUCB-ExpDecay} & $0.399 \pm 0.129$ & $0.417 \pm 0.103$ & $0.560 \pm 0.150$ \\
&\textsc{LinUCB-QG} & $0.670 \pm 0.237$ & $0.619 \pm 0.166$ & $0.867 \pm 0.189$ \\

\hline
\hline
\multirow{4}{*}{0.9}&\textsc{LinUCB} & $0.292 \pm 0.113$ & $0.313 \pm 0.101$ & $0.000 \pm 0.000$ \\
&\textsc{AlwaysQuery} & $0.727 \pm 0.195$ & $\mathbf{0.682 \pm 0.173}$ & $1.000 \pm 0.000$ \\
\cline{2-5}&\textsc{LinUCB-ExpDecay} & $0.399 \pm 0.129$ & $0.408 \pm 0.116$ & $0.560 \pm 0.150$ \\
&\textsc{LinUCB-QG} & $0.670 \pm 0.237$ & $0.644 \pm 0.201$ & $0.867 \pm 0.189$ \\

\hline

\end{tabular}
\caption{Workload data setting $f_{1,sim}$: Querying bandit algorithm metrics in simulation on the test set for different values of $w_{task}$. Averages are across 3 random seeds (For \textsc{LinUCB-ExpDecay}, we also average across 5 policy random seeds). Metric values for \textsc{LinUCB-ExpDecay} and \textsc{LinUCB-QG} correspond to the hyperparameter setting with maximal $M_{wt}$ on the validation set (for each separate data setting).}
\label{table:wtask-0.2-to-0.9-f1}
\end{table*}

\begin{table*}[t]
\centering
\begin{tabular}{ |c|p{1.125 in}||c|c|c||c|c| } 
\hline
\centering $w_{task}$ & \textbf{Method} & $r_{task,avg}$ & $M_{wt}$ & $f_{q}$ \\ 
 \hline
 \hline
\multirow{4}{*}{0.2}&\textsc{LinUCB} & $0.292 \pm 0.113$ & $\mathbf{0.458 \pm 0.023}$ & $0.000 \pm 0.000$ \\
&\textsc{AlwaysQuery} & $0.727 \pm 0.195$ & $-0.255 \pm 0.039$ & $1.000 \pm 0.000$ \\
\cline{2-5}&\textsc{LinUCB-ExpDecay} & $0.303 \pm 0.116$ & $0.438 \pm 0.084$ & $0.160 \pm 0.150$ \\
&\textsc{LinUCB-QG} & $0.336 \pm 0.069$ & $0.353 \pm 0.088$ & $0.333 \pm 0.094$ \\

\hline\hline
\multirow{4}{*}{0.3}&\textsc{LinUCB} & $0.292 \pm 0.113$ & $\mathbf{0.438 \pm 0.034}$ & $0.000 \pm 0.000$ \\
&\textsc{AlwaysQuery} & $0.727 \pm 0.195$ & $-0.132 \pm 0.059$ & $1.000 \pm 0.000$ \\
\cline{2-5}&\textsc{LinUCB-ExpDecay} & $0.303 \pm 0.116$ & $0.421 \pm 0.076$ & $0.160 \pm 0.150$ \\
&\textsc{LinUCB-QG} & $0.336 \pm 0.069$ & $0.351 \pm 0.083$ & $0.333 \pm 0.094$ \\

\hline\hline
\multirow{4}{*}{0.4}&\textsc{LinUCB} & $0.292 \pm 0.113$ & $\mathbf{0.417 \pm 0.045}$ & $0.000 \pm 0.000$ \\
&\textsc{AlwaysQuery} & $0.727 \pm 0.195$ & $-0.009 \pm 0.078$ & $1.000 \pm 0.000$ \\
\cline{2-5}&\textsc{LinUCB-ExpDecay} & $0.303 \pm 0.116$ & $0.404 \pm 0.072$ & $0.160 \pm 0.150$ \\
&\textsc{LinUCB-QG} & $0.336 \pm 0.069$ & $0.349 \pm 0.078$ & $0.333 \pm 0.094$ \\

\hline\hline
\multirow{4}{*}{0.5}&\textsc{LinUCB} & $0.292 \pm 0.113$ & $\mathbf{0.396 \pm 0.056}$ & $0.000 \pm 0.000$ \\
&\textsc{AlwaysQuery} & $0.727 \pm 0.195$ & $0.113 \pm 0.098$ & $1.000 \pm 0.000$ \\
\cline{2-5}&\textsc{LinUCB-ExpDecay} & $0.303 \pm 0.116$ & $0.387 \pm 0.071$ & $0.160 \pm 0.150$ \\
&\textsc{LinUCB-QG} & $0.336 \pm 0.069$ & $0.347 \pm 0.074$ & $0.333 \pm 0.094$ \\

\hline\hline
\multirow{4}{*}{0.6}&\textsc{LinUCB} & $0.292 \pm 0.113$ & $\mathbf{0.375 \pm 0.068}$ & $0.000 \pm 0.000$ \\
&\textsc{AlwaysQuery} & $0.727 \pm 0.195$ & $0.236 \pm 0.117$ & $1.000 \pm 0.000$ \\
\cline{2-5}&\textsc{LinUCB-ExpDecay} & $0.303 \pm 0.116$ & $0.370 \pm 0.075$ & $0.160 \pm 0.150$ \\
&\textsc{LinUCB-QG} & $0.336 \pm 0.069$ & $0.344 \pm 0.071$ & $0.333 \pm 0.094$ \\

\hline\hline
\multirow{4}{*}{0.8}&\textsc{LinUCB} & $0.292 \pm 0.113$ & $0.333 \pm 0.090$ & $0.000 \pm 0.000$ \\
&\textsc{AlwaysQuery} & $0.727 \pm 0.195$ & $\mathbf{0.481 \pm 0.156}$ & $1.000 \pm 0.000$ \\
\cline{2-5}&\textsc{LinUCB-ExpDecay} & $0.303 \pm 0.116$ & $0.336 \pm 0.091$ & $0.160 \pm 0.150$ \\
&\textsc{LinUCB-QG} & $0.560 \pm 0.051$ & $0.466 \pm 0.037$ & $0.800 \pm 0.163$ \\

\hline\hline
\multirow{4}{*}{0.9}&\textsc{LinUCB} & $0.292 \pm 0.113$ & $0.313 \pm 0.101$ & $0.000 \pm 0.000$ \\
&\textsc{AlwaysQuery} & $0.727 \pm 0.195$ & $\mathbf{0.604 \pm 0.176}$ & $1.000 \pm 0.000$ \\
\cline{2-5}&\textsc{LinUCB-ExpDecay} & $0.399 \pm 0.129$ & $0.399 \pm 0.112$ & $0.560 \pm 0.150$ \\
&\textsc{LinUCB-QG} & $0.560 \pm 0.051$ & $0.513 \pm 0.044$ & $0.800 \pm 0.163$ \\

\hline
\end{tabular}
\caption{Workload data setting $f_{2,sim}$: Querying bandit algorithm metrics in simulation on the test set for different values of $w_{task}$. Averages are across 3 random seeds (For \textsc{LinUCB-ExpDecay}, we also average across 5 policy random seeds). Metric values for \textsc{LinUCB-ExpDecay} and \textsc{LinUCB-QG} correspond to the hyperparameter setting with maximal $M_{wt}$ on the validation set (for each separate data setting).}
\label{table:wtask-0.2-to-0.9-f2}
\end{table*}

\begin{table*}[t]
\centering
\begin{tabular}{ |c|p{1.125 in}||c|c|c||c|c| } 
\hline
\centering $w_{task}$ & \textbf{Method} & $r_{task,avg}$ & $M_{wt}$ & $f_{q}$ \\ 
 \hline
 \hline
\multirow{4}{*}{0.2}&\textsc{LinUCB} & $0.292 \pm 0.113$ & $\mathbf{0.429 \pm 0.023}$ & $0.000 \pm 0.000$ \\
&\textsc{AlwaysQuery} & $0.727 \pm 0.195$ & $-0.255 \pm 0.039$ & $1.000 \pm 0.000$ \\
\cline{2-5}&\textsc{LinUCB-ExpDecay} & $0.303 \pm 0.116$ & $0.377 \pm 0.188$ & $0.160 \pm 0.150$ \\
&\textsc{LinUCB-QG} & $0.503 \pm 0.041$ & $0.163 \pm 0.330$ & $0.667 \pm 0.189$ \\

\hline  \hline
\multirow{4}{*}{0.3}&\textsc{LinUCB} & $0.292 \pm 0.113$ & $\mathbf{0.411 \pm 0.034}$ & $0.000 \pm 0.000$ \\
&\textsc{AlwaysQuery} & $0.727 \pm 0.195$ & $-0.132 \pm 0.059$ & $1.000 \pm 0.000$ \\
\cline{2-5}&\textsc{LinUCB-ExpDecay} & $0.303 \pm 0.116$ & $0.367 \pm 0.164$ & $0.160 \pm 0.150$ \\
&\textsc{LinUCB-QG} & $0.503 \pm 0.041$ & $0.206 \pm 0.289$ & $0.667 \pm 0.189$ \\

\hline  \hline
\multirow{4}{*}{0.4}&\textsc{LinUCB} & $0.292 \pm 0.113$ & $\mathbf{0.394 \pm 0.045}$ & $0.000 \pm 0.000$ \\
&\textsc{AlwaysQuery} & $0.727 \pm 0.195$ & $-0.009 \pm 0.078$ & $1.000 \pm 0.000$ \\
\cline{2-5}&\textsc{LinUCB-ExpDecay} & $0.303 \pm 0.116$ & $0.358 \pm 0.142$ & $0.160 \pm 0.150$ \\
&\textsc{LinUCB-QG} & $0.503 \pm 0.041$ & $0.248 \pm 0.249$ & $0.667 \pm 0.189$ \\

\hline  \hline
\multirow{4}{*}{0.5}&\textsc{LinUCB} & $0.292 \pm 0.113$ & $\mathbf{0.377 \pm 0.056}$ & $0.000 \pm 0.000$ \\
&\textsc{AlwaysQuery} & $0.727 \pm 0.195$ & $0.113 \pm 0.098$ & $1.000 \pm 0.000$ \\
\cline{2-5}&\textsc{LinUCB-ExpDecay} & $0.303 \pm 0.116$ & $0.349 \pm 0.122$ & $0.160 \pm 0.150$ \\
&\textsc{LinUCB-QG} & $0.503 \pm 0.041$ & $0.291 \pm 0.209$ & $0.667 \pm 0.189$ \\

\hline  \hline
\multirow{4}{*}{0.6}&\textsc{LinUCB} & $0.292 \pm 0.113$ & $\mathbf{0.360 \pm 0.068}$ & $0.000 \pm 0.000$ \\
&\textsc{AlwaysQuery} & $0.727 \pm 0.195$ & $0.236 \pm 0.117$ & $1.000 \pm 0.000$ \\
\cline{2-5}&\textsc{LinUCB-ExpDecay} & $0.303 \pm 0.116$ & $0.340 \pm 0.106$ & $0.160 \pm 0.150$ \\
&\textsc{LinUCB-QG} & $0.503 \pm 0.041$ & $0.333 \pm 0.170$ & $0.667 \pm 0.189$ \\

\hline  \hline
\multirow{4}{*}{0.8}&\textsc{LinUCB} & $0.292 \pm 0.113$ & $0.326 \pm 0.090$ & $0.000 \pm 0.000$ \\
&\textsc{AlwaysQuery} & $0.727 \pm 0.195$ & $\mathbf{0.481 \pm 0.156}$ & $1.000 \pm 0.000$ \\
\cline{2-5}&\textsc{LinUCB-ExpDecay} & $0.303 \pm 0.116$ & $0.321 \pm 0.096$ & $0.160 \pm 0.150$ \\
&\textsc{LinUCB-QG} & $0.503 \pm 0.041$ & $0.418 \pm 0.093$ & $0.667 \pm 0.189$ \\

\hline  \hline
\multirow{4}{*}{0.9}&\textsc{LinUCB} & $0.292 \pm 0.113$ & $0.309 \pm 0.101$ & $0.000 \pm 0.000$ \\
&\textsc{AlwaysQuery} & $0.727 \pm 0.195$ & $\mathbf{0.604 \pm 0.176}$ & $1.000 \pm 0.000$ \\
\cline{2-5}&\textsc{LinUCB-ExpDecay} & $0.399 \pm 0.129$ & $0.383 \pm 0.111$ & $0.560 \pm 0.150$ \\
&\textsc{LinUCB-QG} & $0.503 \pm 0.041$ & $0.461 \pm 0.059$ & $0.667 \pm 0.189$ \\

\hline
\end{tabular}
\caption{Workload data setting $f_{1,2,sim}$: Querying bandit algorithm metrics in simulation on the test set for different values of $w_{task}$. Averages are across 3 random seeds (For \textsc{LinUCB-ExpDecay}, we also average across 5 policy random seeds). Metric values for \textsc{LinUCB-ExpDecay} and \textsc{LinUCB-QG} correspond to the hyperparameter setting with maximal $M_{wt}$ on the validation set (for each separate data setting).}
\label{table:wtask-0.2-to-0.9-f12}
\end{table*}

\subsubsection{Additional Details: generalization on $D_2$.} 

In this experiment, the validation process is analogous to that for the previous simulation results, but on the test set, we use separate workload models for $\textsc{LinUCB-QG}$ action-selection, and for computing the evaluation metrics $\Delta WL$ and $M_{wt}$. We use either $f_{1,sim}$ or $f_{1,2,sim}$ for the action-selection workload model, while we fix the evaluation workload model to be $f_{2,sim}$, and refer to the two settings as $f_1$-on-$f_2$ and $f_{1,2}$-on-$f_2$, respectively. Table \ref{tab:bandit-cross-comparison} shows the results for $\textsc{LinUCB-QG}$ in each of the two settings.

\begin{table}[ht]
    \centering
    \scriptsize
    \begin{tabular}{|c|c|c|}
    \hline
        \textbf{Metric} & \textbf{$f_1$-on-$f_2$} & \textbf{$f_{1,2}$-on-$f_2$} \\ 
        \hline
        \hline
        $r_{task,avg}$      & $0.434 \pm 0.049$  & $0.503 \pm 0.041$    \\ 
        $\Delta WL$    & $-0.159 \pm 0.092$  & $-0.286 \pm 0.159$   \\ 
        $M_{wt}$       & $0.269 \pm 0.040$   & $\mathbf{0.373 \pm 0.108}$    \\ 
        $f_q$     & $0.733 \pm 0.094$   & $0.667 \pm 0.189$    \\ 
        \hline
    \end{tabular}
    \caption{\textsc{LinUCB-QG} performance in cross-data setting at test time, showing algorithm metrics in simulation on the test set, with $w_{task} = 0.4$. Averages are across 3 random seeds. Values for $r_{task,avg}$ and $\Delta WL$ correspond to the hyperparameter setting with maximal $M_{wt}$ on the validation set (for each separate data setting).}
    \label{tab:bandit-cross-comparison}
\end{table}

As mentioned in Section \ref{subsec:simulated-exps}, we find that the mean weighted metric $M_{wt}$ is higher in the $f_{1,2}$-on-$f_2$ than the $f_1$-on-$f_2$ setting (and $\Delta WL$ and $f_q$ are both lower). 

\subsubsection{Sample Workload Model Rollout}

Finally, Figure \ref{fig:appendix-gap} shows a rollout of the workload model in simulation for \textsc{LinUCB-QG}, showing how the workload $WL_t$ evolves over time, how the UCB estimate gap variable $G$ varies, and when \textsc{LinUCB-QG} decides to query. In this example, \textsc{LinUCB-QG} decides to ask the human for help when $G > w WL_t$ (in this rollout, $w = 4$), but also recall that once \textsc{LinUCB-QG} has asked for help for the current food item, it will continue to execute the expert action until it has converged.

\subsection{Experiments: Real-world User Study}

\subsubsection{Feeding Tool}

In our work, we use a custom feeding tool developed in prior work [\citeconsecutive{rajatpriyarss2024, belkhale2022balancing}], which is attached to the end of the Kinova's Robotiq 2F-85 gripper. The tool consists of a motorized fork with two degrees of freedom: pitch $\beta$ and roll $\gamma$. Each of the six robot actions $a_r$ (for $r \in \{1, \dots, 6\}$) corresponds to a unique value of $\beta$ and $\gamma$. In particular, the pitch configuration (either TA, VS, or TV) affects the value of $\beta$, while the roll configuration (either $0^\circ$ or $90^\circ$) affects the value of $\gamma$.

\subsubsection{Details on Perception Modules} To more accurately determine skewering-relevant geometric information for each bite-sized food item, we leverage perception modules that were developed in prior work \cite{rajatpriyarss2024}. We use GroundingDINO \cite{liu2023grounding} to produce an initial bounding box around the food item, followed by SAM \cite{kirillov2023segment} to segment the food item, and finally extract the minimum area rectangle around the segmentation mask to get a refined bounding box. From this bounding box, we derive the centroid and major axis of the food item, which are used by the low-level action executor to execute the desired robot action $a_r$.

\subsubsection{Additional Details and Analysis}

Here we present additional details related to the real-world user study experiments outlined in Section \ref{subsec:kinova-study}.

\textit{Pre-method questions.} We ask the users with mobility limitations the following questions prior to each method:
\begin{itemize}[leftmargin=*,itemsep=-0.1em]
    \item ``How mentally/physically burdened do you feel currently?"
    \item ``How hurried or rushed do you feel currently?"	
\end{itemize}

For users without mobility limitations, we ask the following questions:

\begin{itemize}[leftmargin=*,itemsep=-0.1em]
    \item ``How mentally burdened do you feel currently?"	
    \item ``How hurried or rushed do you feel currently?"	
\end{itemize}

\textit{Post-method questions.} We ask the users with mobility limitations the following questions after each method:
\begin{itemize}[leftmargin=*,itemsep=-0.1em]
    \item ``For the last method, how mentally/physically burdened do you feel currently because of the robot querying you?" 	
    \item ``For the last method, how hurried or rushed do you feel currently because of the robot querying you?"
    \item ``For the last method, how hard did you have to work to make the robot pick up food items?"
    \item ``For the last method, how successful was the robot in picking up food items?"
\end{itemize}

For users without mobility limitations, we ask the following questions:
\begin{itemize}[leftmargin=*,itemsep=-0.1em]
    \item ``For the last method, how mentally burdened do you feel currently because of the robot querying you?" 	
    \item ``For the last method, how hurried or rushed do you feel currently because of the robot querying you?"
    \item ``For the last method, how hard did you have to work to make the robot feed you?"
    \item ``For the last method, how successful was the robot in picking up the food item and bringing it to your mouth?"
\end{itemize}

\textit{Compensation.} Users were compensated at a rate of \$12/hr, for a total compensation of \$18 for the entire study (as the study duration was 1.5 hr).

\textit{Note on initial workload $WL_0$.} Although we do ask for an initial estimate of the user's workload by asking the pre-method questions above, in our real-world experiments, we provide a fixed initial workload value of $WL_0 = 0.5$ for \textsc{LinUCB-QG}. This value is used by \textsc{LinUCB-QG} for its internal workload predictions and decision-making. However, when calculating the subjective metrics in Section \ref{subsec:kinova-study} that measure changes in workload (Delta-Mental, Delta-Temporal, and Delta-Querying Workload), we use the responses to the pre-method questions to estimate the user's initial workload.

\textit{Method comparison across all metrics.} Tables \ref{table:userstudy-subjective-table} and \ref{table:userstudy-objective-table} include comparisons across the three methods used in the real-world user study (\textsc{LinUCB}, \textsc{AlwaysQuery}, \textsc{LinUCB-QG}) for the 7 subjective and 3 objective metrics mentioned in Section \ref{subsec:kinova-study}, respectively. For all metrics, we find that \textsc{LinUCB-QG} achieves an intermediate value for that metric compared to the other two algorithms, suggesting that our method finds a balance between querying workload and task performance (which these metrics cover). Additionally, Table \ref{table:userstudy-pvals} shows the full results of the Wilcoxon paired signed-rank tests that we used to determine whether \textsc{LinUCB-QG} exhibited statistically significant differences in the full set of metrics compared to \textsc{LinUCB-QG} and \textsc{AlwaysQuery}. For the metrics related to querying workload, our alternative hypothesis was that the value of the metric for \textsc{LinUCB-QG} was less than the value for \textsc{AlwaysQuery}. For the metrics related to task performance, our alternative hypothesis was that the value of the metric for \textsc{LinUCB-QG} was greater than the value for \textsc{LinUCB}. For all metrics, we find that \textsc{LinUCB-QG} has a statistically significant improvement compared to the baseline for that metric.

\begin{table*}[t]
\centering
\begin{tabular}{ |p{1 in}||c|c|c|c|c|c|c| } 
\hline
\centering \textbf{Method} & Post-Mental &  Post-Temporal & Post-Performance & Post-Effort & Delta-Mental & Delta-Temporal & Delta-Querying Workload \\ 
 \hline
 \hline
 \textsc{LinUCB} & $1.24 \pm 0.59$ & $1.37 \pm 0.97$ & $3.03 \pm 0.77$ & $1.24 \pm 0.75$ & $-0.34 \pm 0.88$ & $-0.26 \pm 0.92$ & $-0.02 \pm 0.16$ \\
\textsc{AlwaysQuery} & $3.03 \pm 1.38$ & $2.58 \pm 1.55$ & $4.67 \pm 0.72$ & $3.03 \pm 1.33$ & $1.42 \pm 1.39$ & $0.87 \pm 1.76$ & $0.39 \pm 0.31$ \\
\textsc{LinUCB-QG} & $2.26 \pm 1.11$ & $2.08 \pm 1.22$ & $4.06 \pm 1.04$ & $2.24 \pm 1.05$ & $0.63 \pm 1.05$ & $0.37 \pm 1.08$ & $0.21 \pm 0.20$ \\
\hline
\end{tabular}
\caption{Subjective metric results for the real-world user study.}
\label{table:userstudy-subjective-table}
\end{table*}

\begin{table*}[t]
\centering
\begin{tabular}{ |p{1.125 in}||c|c|c| } 
\hline
\centering \textbf{Method} & $r_{task,avg}$ &  $n_{success}$ & $n_{q}$ \\ 
 \hline
 \hline
\textsc{LinUCB} & $0.38 \pm 0.08$ & $2.00 \pm 0.23$ & - \\
\textsc{AlwaysQuery} & $0.87 \pm 0.25$ & $2.68 \pm 0.61$ & $3 \pm 0$ \\
\textsc{LinUCB-QG} & $0.68 \pm 0.29$ & $2.53 \pm 0.72$ & $1.74 \pm 0.78$ \\
\hline
\end{tabular}
\caption{Objective metric results for the real-world user study.}
\label{table:userstudy-objective-table}
\end{table*}

\begin{table*}[t]
\centering
\begin{tabular}{ |p{1.25 in}|c|c| } 
\hline
\centering \textbf{Metric} & Hypothesis & $p$\\ 
 \hline
 \hline
Post-Mental & \multirow{7}{*}{\textsc{LinUCB-QG} $<$ \textsc{AlwaysQuery} } & $3.28\text{e-}4$ \\   
Post-Temporal && $1.42\text{e-}2$ \\           
Post-Effort && $1.2\text{e-}4$  \\             
Delta-Mental && $7.68\text{e-}4$ \\            
Delta-Temporal && $1.82\text{e-}2$ \\          
Delta-Querying Workload && $1.47\text{e-}4$ \\  
$n_q$ && $2.93\text{e-}7$ \\  
\hline
Post-Performance & \multirow{3}{*}{\textsc{LinUCB-QG} $>$ \textsc{LinUCB} } & $1.59\text{e-}4$ \\  
$r_{task,avg}$ && $1.56\text{e-}4$ \\ 
$n_{success}$ && $2.18\text{e-}4$ \\  
\hline
\end{tabular}
\caption{Wilcoxon paired signed-rank test results for the real-world user study, showing $p$-values associated with querying workload metrics (top) and task performance metrics (bottom). For all tests, we used a significance level of $\alpha = 0.05$.}
\label{table:userstudy-pvals}
\end{table*}

\textit{Comments on Post-Performance metric.} When looking at the Post-Performance metric, we find that \textsc{LinUCB-QG} achieves an intermediate value, higher than \textsc{LinUCB} but lower than \textsc{Always-Query}. When asking users to rate subjective success, we asked them to rate the overall success of each algorithm, which depends on both whether the algorithm successfully picked up food items and how many attempts were required for success. For this reason, we believe that users rated the success of \textsc{Always-Query} highly because it efficiently selects the optimal action for each food item within one timestep, whereas \textsc{LinUCB-QG} sometimes takes multiple timesteps if it does not query for a particular food item. 

\section*{ACKNOWLEDGMENT}

This work was partly funded by NSF CCF 2312774 and NSF OAC-2311521, a LinkedIn Research Award, and a gift from Wayfair, and by NSF IIS 2132846 and CAREER 2238792. Research reported in this publication was additionally supported by the Eunice Kennedy Shriver National Institute Of Child Health \& Human Development of the National Institutes of Health and the Office of the Director of the National Institutes of Health under Award Number T32HD113301. The content is solely the responsibility of the authors and does not necessarily represent the official views of the National Institutes of Health.

The authors would like to thank Ethan Gordon for his assistance with the food dataset, Ziang Liu, Pranav Thakkar and Rishabh Madan for their help with running the robot user study, Janna Lin for providing the voice interface, Shuaixing Chen for helping with figure creation, Tom Silver for paper feedback, and all of the participants in our two user studies.

\end{document}